\documentclass[journal,trans]{IEEEtran}

\usepackage[usenames,dvipsnames]{xcolor}

\ifCLASSINFOpdf
\else
\fi

\usepackage[commandnameprefix=always]{changes}
\definechangesauthor[name=sg, color=orange]{SG}

\usepackage{etex}
\usepackage{algorithmicx}
\usepackage{algpseudocode}
\usepackage[ruled,vlined]{algorithm2e}
\SetKwRepeat{Do}{do}{while}

\usepackage{balance}
\usepackage[justification=centering]{caption}
\usepackage[noadjust]{cite}
\usepackage{array}
\usepackage{graphicx}
\usepackage{epstopdf}
\usepackage{setspace}
\usepackage[export]{adjustbox}
\usepackage{ctable}
\usepackage{enumerate}
\usepackage{ragged2e}
\usepackage{multirow}
\usepackage{tabularx,tabulary}
\usepackage{nicefrac}
\usepackage{siunitx}
\usepackage[caption=false,font=footnotesize,labelformat=simple]{subfig} 
\usepackage[flushleft]{threeparttable}

\newlength{\textfloatsepsave}
\usepackage{
  tikz,
  pgfplots,
  pgfplotstable
}
\usepackage{hyperref}
\usepackage{pifont}
\newcommand{\cmark}{\ding{51}}%
\newcommand{\xmark}{\ding{55}}%
\newcommand{\circleone}{\ding{202}}
\newcommand{\circletwo}{\ding{203}}
\newcommand{\circlethree}{\ding{204}}

\usepackage{rotating}
\usepackage{tabularx,ragged2e,booktabs}
\newcolumntype{C}[1]{>{\Centering}m{#1}}

\usepackage{mathtools}
\usepackage{amsmath,amssymb,amsthm}

\usepackage{authblk}
\usepackage{verbatim}

\usepackage{tcolorbox}

\usepackage[final]{pdfpages}
\usepackage{pifont}
\usepackage{framed}
\usepackage{soul}
\usepackage{booktabs}
\usepackage{multirow}
\usepackage{mathtools}

\usetikzlibrary{
  shapes.geometric,
  arrows,
  external,
  pgfplots.groupplots,
  matrix
}

\pgfplotsset{compat=1.9}

\usepackage{mathtools}

\DeclarePairedDelimiter\norm{\lVert}{\rVert}

\DeclareMathAlphabet{\mathcal}{OMS}{cmsy}{m}{n}
\DeclareGraphicsExtensions{%
    .png,.PNG,%
    .pdf,.PDF,%
    .jpg,.mps,.jpeg,.jbig2,.jb2,.JPG,.JPEG,.JBIG2,.JB2}
\usepackage{xparse}
\newcommand{\bnm}{\begin{newmath}}
\newcommand{\enm}{\end{newmath}}

\newcommand{\bea}{\begin{eqnarray*}}%
\newcommand{\eea}{\end{eqnarray*}}%

\newcommand{\bne}{\begin{newequation}}
\newcommand{\ene}{\end{newequation}}

\newcommand{\bal}{\begin{newalign}}
\newcommand{\eal}{\end{newalign}}

\newenvironment{newalign}{\begin{align}%
\setlength{\abovedisplayskip}{4pt}%
\setlength{\belowdisplayskip}{4pt}%
\setlength{\abovedisplayshortskip}{6pt}%
\setlength{\belowdisplayshortskip}{6pt} }{\end{align}}

\newenvironment{newmath}{\begin{displaymath}%
\setlength{\abovedisplayskip}{4pt}%
\setlength{\belowdisplayskip}{4pt}%
\setlength{\abovedisplayshortskip}{6pt}%
\setlength{\belowdisplayshortskip}{6pt} }{\end{displaymath}}

\newenvironment{newequation}{\begin{equation}%
\setlength{\abovedisplayskip}{4pt}%
\setlength{\belowdisplayskip}{4pt}%
\setlength{\abovedisplayshortskip}{6pt}%
\setlength{\belowdisplayshortskip}{6pt} }{\end{equation}}

\newcounter{ctr}

\newcounter{mytable}
\def\mytable{\begin{centering}\refstepcounter{mytable}}
\def\endmytable{\end{centering}}

\newcounter{myfig}
\def\myfig{\begin{centering}\refstepcounter{myfig}}
\def\endmyfig{\end{centering}}

\newlength{\saveparindent}
\newlength{\saveparskip}
\newcommand{\E}{{\rm I\kern-.3em E}}

\renewcommand{\eqref}[1]{\mbox{Equation~(\ref{#1})}}

\def \part {part}

\DeclareMathOperator*{\argmin}{argmin}

\renewcommand{\paragraph}[1]{\vspace*{6pt}\noindent\textbf{#1}\;}

\def \blackslug{\hbox{\hskip 1pt \vrule width 4pt height 8pt
    depth 1.5pt \hskip 1pt}}
\def \qed{\quad\blackslug\lower 8.5pt\null\par}

\newcounter{mynote}[section]

\newcommand\ignore[1]{}

\newcounter{rcnote}[section]

\newcounter{mrnote}[section]

\newcounter{fknote}[section]

\newcounter{anote}[section]

\DeclareMathSymbol{\mlq}{\mathord}{operators}{``}
\DeclareMathSymbol{\mrq}{\mathord}{operators}{`'}

\newcommand{\rhf}[2]{R_{f, \gamma}}

\DeclareDocumentCommand{\edist}{o o}{
  \ensuremath{
    \IfNoValueTF{#1}{{d}}{{\sf d}(#1,#2)}
  }
}

\newcommand{\olrk}[1]{\ifx\nursymbol#1\else\!\!\mskip4.5mu plus 0.5mu\left(\mskip0.5mu plus0.5mu #1\mskip1.5mu plus0.5mu \right)\fi}

\NewDocumentCommand{\indseq}{ O{1} O{r} }{{#1}\ldots {#2}}

\newcommand{\specialcell}[2][c]{%
  \begin{tabular}[#1]{@{}c@{}}#2\end{tabular}}

\newcommand{\solution}{Bulls-Eye}
\newcommand{\btbf}[1]{{\color{blue}{\textbf{#1}}}}
\newcommand{\rtbf}[1]{{\color{red}{\textbf{#1}}}}

\usepackage{acronym}
\acrodef{IC}{integrated circuit}
\acrodef{EDA}{electronic design automation}
\acrodef{HDL}{hardware description language}
\acrodef{AIG}{and-inverter-graph}
\acrodef{RL}{reinforcement learning}
\acrodef{ML}{machine learning}
\acrodef{IP}{intellectual property}
\acrodef{RTL}{register transfer level}
\acrodef{DAG}{directed acyclic graph}
\acrodef{GCN}{graph convolutional network}
\acrodef{QoR}{quality-of-result}
\acrodef{SoC}{system-on-chip}
\acrodef{MIG}{Majority inverter graph}
\acrodef{XAG}{XOR AND graph}
\acrodef{DNN}{deep neural network}
\acrodef{MDP}{Markov decision process}
\acrodef{CNN}{convolutional neural network}
\acrodef{RL}{reinforcement learning}
\acrodef{MSE}{mean-square error}
\acrodef{SA}{simulated annealing}
\acrodef{SOTA}{state-of-the-art}

\begin{document}


\title{Too Big to Fail? Active Few-shot Learning Guided Logic Synthesis}


\author{\IEEEauthorblockN{Animesh Basak Chowdhury\IEEEauthorrefmark{1},
Benjamin Tan\IEEEauthorrefmark{2},
Ryan Carey\IEEEauthorrefmark{3}, 
Tushit Jain\IEEEauthorrefmark{3},
Ramesh Karri\IEEEauthorrefmark{1} and
Siddharth Garg\IEEEauthorrefmark{1}}

\IEEEauthorblockA{\IEEEauthorrefmark{1}Dept. of Electrical and Computer Engineering, Tandon School of Engineering, New York University}

\IEEEauthorblockA{\IEEEauthorrefmark{2}Dept. of Electrical and Software Engineering, Schulich School of Engineering, University of Calgary}

\IEEEauthorblockA{\IEEEauthorrefmark{3}Qualcomm Technologies, Inc., San Diego, CA 92121 USA}
}





\maketitle

\begin{abstract}
Generating sub-optimal synthesis transformation sequences (``synthesis recipe") is an important problem in logic synthesis. Manually crafted synthesis recipes have poor quality. State-of-the art \ac{ML} works to generate synthesis recipes do not scale to large netlists as the models need to be trained from scratch, for which training data is collected using time consuming synthesis runs. We propose a new approach, Bulls-Eye, that fine-tunes a pre-trained model on past synthesis data to accurately predict the quality of a synthesis recipe for an unseen netlist. This approach on achieves 2x-10x run-time improvement and better \ac{QoR} than state-of-the-art machine learning approaches.
\end{abstract}


\IEEEpeerreviewmaketitle

\section{Introduction}
\label{sec:intro}

There is a growing academic and industry interest in using \acf{ML} techniques in design automation problems~\cite{googlerl,lsoracle,lin2020dreamplace}. 
Several problems in \ac{EDA}, such as logic synthesis, placement and routing, and VLSI testing are combinatorial optimization problems that require sequential decision-making to achieve the target objective. Logic synthesis, the first step in an \ac{EDA} flow, applies a sequence of synthesis transformations (i.e., a ``synthesis recipe'') to an \ac{AIG} representation of a Boolean function to minimize its size or depth.
For instance, the transformation heuristics in ABC~\cite{abc}, a leading academic synthesis tool, include steps like \texttt{balance}, \texttt{refactor}, \texttt{rewrite}, etc.
In this work, we focus on ML-guided search for ``good'' logic synthesis recipes; a problem that has received much recent attention.


Traditionally, EDA engineers choose from a set 
handcrafted synthesis recipes, for instance, \texttt{resyn2} or \texttt{compress2rs}~\cite{abc} 
based on intuition and experience. 
Yet, the space of all synthesis recipes is \emph{vast}; recent work has shown 
\ac{ML}-guided exploration of this vast solution space can yield synthesis recipes that are 
tailored for each design and outperform handcrafted recipes in terms of \acf{QoR}~\cite{cunxi_dac,drills,mlcad_abc,iccad_2021}.
In these papers, finding a good synthesis recipe is formulated as a classification problem~\cite{cunxi_dac}---which seeks to distinguish good ``angel'' recipes from bad ``devil'' recipes---or as in~\cite{drills,mlcad_abc,iccad_2021} as an \ac{MDP} that is solved using \acf{RL}. 
However, the prior work is deficient on two fronts: (1) scalability to large designs; and (2) generalizability to previously unseen netlists. 
Typically, models are trained \textbf{from scratch}, requiring significant training time and a large number of synthesis runs for large unseen designs; and leaving prior experience \textbf{untapped}. 
For this reason, in [12], ``benchmarks were excluded from the experimentation which showed significant training times, being large circuits''. 

The underlying issue is that ML methods rely on high-quality labeled data (or reward signals for RL)---in logic synthesis, the input features are AIGs and synthesis recipes, and the resulting QoRs are labels/rewards. For each new design, prior work obtains labels/rewards (i.e., QoRs) by synthesizing the design using a large number of synthesis recipes; however, this is prohibitively time-consuming for large designs. An alternative is to use fewer synthesis runs, or even a model trained on previously seen designs, but this results in lower quality synthesis recipes.  

In this paper, we propose a new logic ML-guided synthesis solution that addresses these limitations and scales to \emph{large} designs on which existing \ac{SOTA} solutions fail or time-out.  
\solution{} has two key ideas: (1) few-shot learning: fine-tuning a base model trained on previously seen designs using only a few training samples from the new design;  and 
(2) active learning: fine-tuning can be further improved by intelligently picking the synthesis recipes for which we obtain QoRs/labels given a limited synthesis run budget. 
We illustrate these ideas using \textbf{\solution{}}, a framework that uses a \ac{GCN} as a \ac{QoR} predictor, and \acf{SA} to search the design space. The core ideas proposed are general and apply to other ML-guided synthesis frameworks also. 
%
To summarize, our key contributions are:
\begin{itemize}
    \item \textbf{\solution{}}: an active few-shot learning approach that transfers knowledge from a base model trained on past data to train a new model that predicts the \ac{QoR} of a synthesis run for an unseen netlist; and uses simulated annealing to find an optimized synthesis recipe for the new netlist.
    \item Scalability to large benchmarks: Bulls-Eye's base model can be learned on small designs for which training data is easy to generate; and transferred using only a few, smartly selected synthesis runs (i.e., "shots") to large, unseen netlists. Bulls-Eye produces high-quality synthesis recipes for benchmarks on which SOTA methods fail or time-out.
    \item \solution{} achieves better \ac{QoR} with \textbf{2x-10x} run-time speed-up and more than \textbf{7x} speed-up on large benchmarks compared to conventional approaches. Compared to prior \ac{SOTA}, \solution{} achieves better \ac{QoR} with an average \textbf{1.04x} ( vs. \cite{iccad_2021}) and \textbf{11.21x} (vs. \cite{mlcad_abc}) run-time speedup.   
\end{itemize}

\section{Preliminaries and Related work}
\label{sec:bg}

\subsection{Logic Synthesis}
\label{subsec:logic_synthesis}
Logic synthesis transforms a hardware design in a high-level abstraction (e.g., register-transfer level) to a gate-level netlist, ultimately mapping to a user-specified technology library.
Typically, logic synthesis tries to minimize the circuit area and meet a specific delay constraint. 
Logic synthesis is performed in a series of steps: (1) the design is converted to a generic netlist-level representation (e.g., \ac{AIG}), (2) the netlist undergoes heuristics-based optimization to implement the same functionality using fewer gates and/or reduced depth which correlates to a reduction in the final area and delay of design~\cite{lsoracle}, (3) technology mapping of the generic netlist using standard cells, and (4) post-mapping optimizations. Consistent with prior works, we focus on the pre-technology mapping optimization of generic gate-level netlists.

\begin{table}[tb]
\caption{Work on \ac{ML}-based synthesis recipe generation} 
\centering
\setlength\tabcolsep{4pt}
\resizebox{\columnwidth}{!}{
\begin{tabular}{@{}ccccc@{}}
\toprule
\multirow{2}{*}{\begin{tabular}[c]{@{}c@{}}{Prior} \\ {work}\end{tabular}} &  \multirow{2}{*}{\begin{tabular}[c]{@{}c@{}}{Input}\\ {Features}\end{tabular}} & \multirow{2}{*}{\begin{tabular}[c]{@{}c@{}}{ML}\\ {techniques}\end{tabular}} & \multirow{2}{*}{\begin{tabular}[c]{@{}c@{}}{Max. nodes}\\ {considered}\end{tabular}} & \multirow{2}{*}{\begin{tabular}[c]{@{}c@{}}{Model}\\ {transferability}\end{tabular}} \\
 & & & & \\ \midrule
\cite{firstWorkDL_synth} & MIG & GCN, RL & $\le$2000 & \xmark \\
\cite{cunxi_dac} & Flow encoding & CNN & 44045 & \xmark\\
\cite{drills} & AIG encoding & RL & 176938 & $\dagger$\\
\cite{cunxi_iccad} & - & MAB & 30003 & \xmark \\ 
\cite{mlcad_abc} & AIG+state encoding & GCN, RL & 2675 & \xmark \\
\cite{iccad_2021} & AIG+state encoding & GCN,RL & 32060 & \xmark\\
\textbf{\textit{\solution}} & AIG+recipe encoding & GCN,CNN & 114771 & \cmark \\
\bottomrule
\end{tabular}
}
\small{$\dagger$ \textit{DRiLLS claims the model can be reused, without empirical evaluation}}
\label{table:priorWork}
\vspace{-2em}
\end{table}

ABC~\cite{abc} is the current state-of-the-art framework for optimizing \ac{AIG}-based netlists. 
\acp{AIG} are \acp{DAG} 
and ABC's synthesis transformation heuristics perform local sub-graph-level optimization to reduce the \ac{AIG} structure.
ABC's well-known transformations are \texttt{refactor, rewrite, re-substitute} (these reduce the nodes), and balance (for depth-reduction). 
For $\mathbf{M}$ unique synthesis transformations, $\{\mathbf{T_{1}}, \mathbf{T_{2}} ... \mathbf{T_M}\}$ in a synthesis recipe $\mathbf{S}$, the possible number of synthesis recipes of length $\mathbf{L}$ is  $\mathbf{M}^\mathbf{L}$ (with repeatable transformation).
This indicates that the search space of possible synthesis recipes is huge. 
The problem of optimal synthesis recipe generation for a given \ac{AIG} is: 
\begin{align}
\argmin_{\mathbf{S}} QoR(\mathcal{G} (\mathbf{AIG},\mathbf{S}))
\end{align}
where $\mathcal{G}$ is the synthesis function defined as $\mathcal{G} : AIG \times S \longrightarrow AIG$ and $QoR$ is a function that evaluates the \textit{quality} of the synthesis (i.e., lower number of nodes/depth of \ac{AIG} is better).

\subsection{Prior work}
We summarize prior work on \ac{ML} for logic synthesis in~\autoref{table:priorWork}. 
Recent enhancements in logic synthesis have built on recent successes of supervised learning and \ac{RL} approaches for solving sequential decision making problems~\cite{silver2017mastering,googlerl} in an attempt to find good synthesis recipes in the vast space of possible recipes. 
In \cite{firstWorkDL_synth,drills,mlcad_abc,iccad_2021}, authors generate optimal synthesis by formulating the problem as an \ac{MDP} and training a policy gradient agent over a collection of synthesis runs on a set of benchmarks. 
During inference, the trained agent (in exploit phase) predict the best synthesis recipe for a given \ac{AIG}. 
Authors in \cite{firstWorkDL_synth,drills} use handcrafted features from \acp{AIG} (i.e. number of AND and NOT gates, depth, number of primary inputs and outputs) to represent \ac{AIG} state. 
Zhu \textit{et al.} use handcrafted features with \ac{GCN} embedding of \ac{AIG} to represent current \ac{AIG} state~\cite{mlcad_abc}. 
These works show results that improve upon handcrafted recipes like \texttt{resyn2}.
%
%
In an alternate approach, Yu \textit{et al.} classify an unseen synthesis recipe as good (``angel'') or bad (``devil'') by training a model on the results corresponding to a set of recipes for a  design~\cite{cunxi_dac}. 
However, the trained model is design specific; a model trained on one design cannot be reused for another.

\section{Pitfalls of existing approaches}
\label{sec:motiv}

Black-box optimization techniques (e.g., simulated annealing or \ac{RL} methods in prior work) need feedback in terms of the \textit{cost} (or \textit{reward}) of  actions; in the logic synthesis context, the action is a synthesis recipe $\mathbf{S}$ and the cost/reward is $\mathcal{G} (\mathbf{AIG},\mathbf{S})$, obtained by synthesizing $\mathbf{AIG}$ using $\mathbf{S}$. Typically a large number of cost/reward evaluations, or equivalently, synthesis runs, are required to obtain good solutions.  
The biggest challenge here is that even a single synthesis run for 
modern complex \ac{SoC} designs 
can require several hours or days, which limits the number of synthesis runs that can be performed in a limited time budget, and consequently, the amount of training data that can be collected and the fraction of design space that can explored.  
A black-box optimizer can cover in a limited time.    
Hence, although \ac{RL}-based approaches have shown promise in generating good quality synthesis recipes, they cannot scale up to industrial-sized benchmarks~\cite{mlcad_abc,iccad_2021}. 
These observations indicate the following key questions computationally-efficient \ac{ML}-enhanced logic synthesis:
\begin{itemize}
    \item How can we predict the best synthesis recipe for a new, unseen netlist using only a limited or budgeted number of synthesis runs of that netlist? 
    \item  Given a budget, which synthesis recipes should be selected to train an accurate model for the new netlist?
    \item How can we efficiently leverage past information, specifically, synthesis runs on previously seen netlists to improve solution quality?
\end{itemize}
We next present \solution{}, that solves the key roadblocks towards computationally efficient \ac{ML}-guided logic synthesis.
\section{The Bullseye framework}
\label{sec:framework}

To reduce the cost of running actual synthesis during black-box optimization, \solution{} trains a \textit{QoR} predictor, $\mathbf{\hat{F}} (\mathbf{AIG},\mathbf{S}, \mathbf{\theta})$, which predicts the \textit{QoR} for a given 
$\mathbf{AIG}$ synthesized using an $L$-length synthesis recipe $\mathbf{S}$. 
We now seek to solve the following problem:
 \begin{align}
 \argmin_{\mathbf{S}} \mathbf{\hat{F}} (\mathbf{AIG},\mathbf{S}, \mathbf{\theta}) \label{eq:proxyMod} \\
    \mathbf{\hat{F}} \approx QoR(\mathcal{G}(\mathbf{AIG},\mathbf{S})) \nonumber
 \end{align}
where $\mathbf{\theta}$ represents the QoR predictor's model parameters, and $\epsilon$ represents model noise that we also seek to minimize.

\subsection{Insights from \ac{ML} domain}
Unlike \ac{ML} application in computer vision, the diversity of circuit characterization in terms of size, complexity, functionalities is not well studied and understood. We borrow two important insights from core \ac{ML} community to solve our problem of learning proxy model with limited data.

\textbf{\textit{Insight 1}}: In computer vision, various deep learning networks learn robust embeddings and are pre-trained on large dataset like Imagenet~\cite{krizhevsky2012imagenet}. The pre-trained models are fine-tuned later for specific applications.

This helps us adopt a similar strategy and study its applicability in \ac{EDA}. Large circuit designs have smaller known IP designs as building blocks; therefore pre-training a model using labelled data on small sized IP designs followed by finetuning can help predict important parameters of downstream tasks. Thus, a one-time training with labelled data on small IP designs (varied functionality) can be a qualitative QoR predictor.

\textbf{\textit{Insight 2}}: Few-shot learning~\cite{wang2020generalizing} learns a hypothesis function with limited labeled data of target task; however abundant labeled data is available for different tasks. Effectively, it uses transfer learning approach for classification/regression task using limited labeled data of target task.

This insight guide us in our problem context for solving the problem of data scarcity, knowledge transfer and picking best-possible synthesis scripts (few-shots) to generate the labelled data. We solve this problem in three steps, illustrated in \noindent \autoref{fig:overall}.
\circleone{} First we train a \emph{zero-shot} \ac{QoR} predictor using prior synthesis data; as we show, it is sufficient to use \emph{small} benchmarks to train this zero-shot model.   
\circletwo{} Next we fine-tune the zero-shot model using a small number of smartly selected synthesis runs of the AIG being optimized; and \circlethree{} we deploy the fine-tuned \ac{QoR} predictor within a black-box SA optimizer.

\subsection{Zero-Shot \ac{QoR} predictor (\circleone{})}

We assume that a design house has access to data from previously synthesized designs or can quickly generate data by running synthesis on \emph{small} designs. We also developed OpenABC-D~\cite{bchowdhury2021openabc}), an open-source automated synthesis data generation framework used to generate data and train our Zero-shot predictor.

In this work, we adopt a \ac{GCN}-based architecture for the predictor (\autoref{fig:OpenABCDnetwork}); GCNs have been used in RL-based approaches as well~\cite{iccad_2021,mlcad_abc}, but with different architectures. Our predictor architecture is shown in ~\autoref{fig:OpenABCDnetwork}a and contains two parallel paths. 
The first path takes an AIG as input and create an AIG embedding of the network; the second path extract features from synthesis recipe and outputs a ``synthesis recipe embedding". We now describe the ``AIG embedding" network and recipe embedding network:

\begin{figure}
    \centering
     \subfloat[][Network architecture]{\includegraphics[scale=0.8]{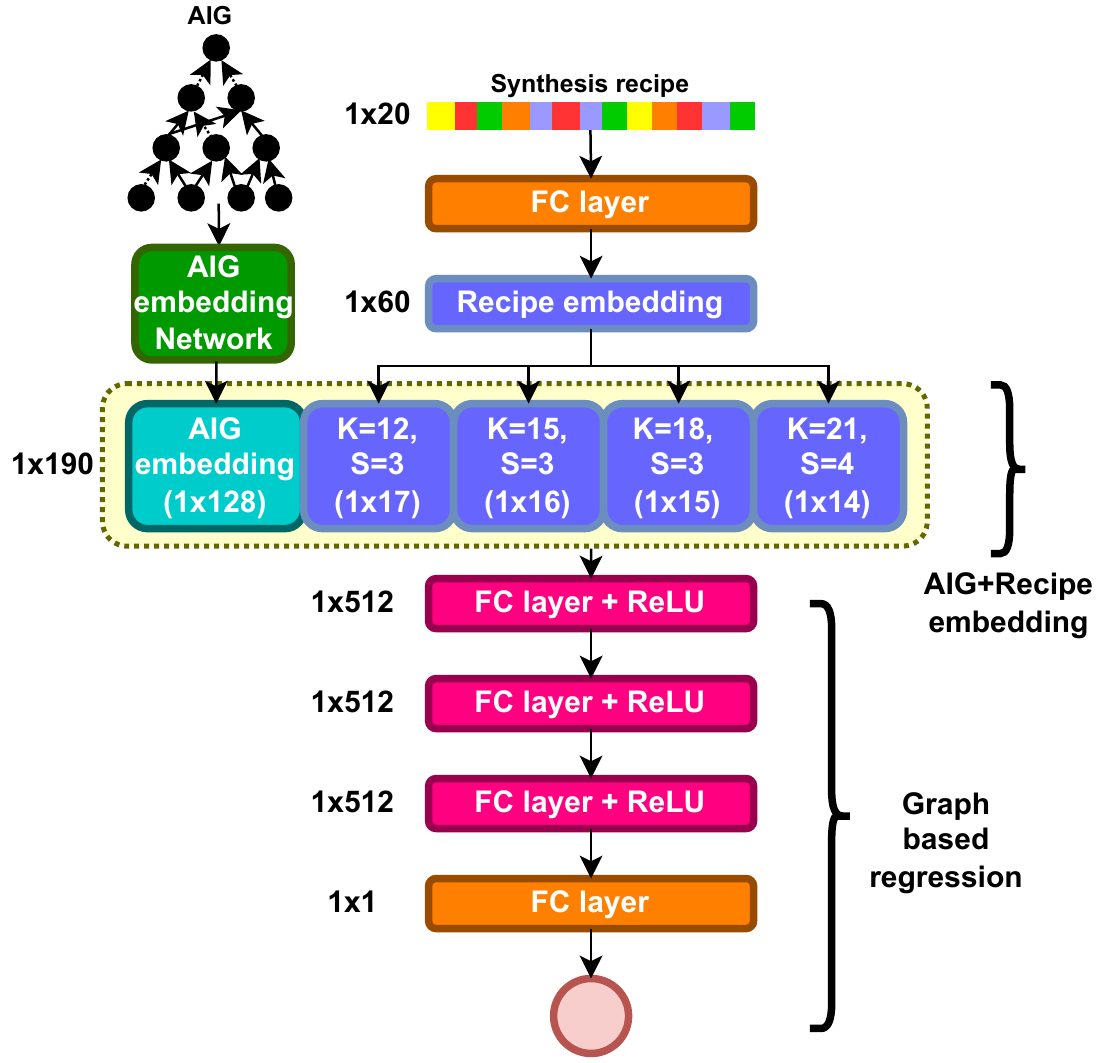}} \\
     \subfloat[][AIG embedding Network]{\includegraphics[width=\columnwidth]{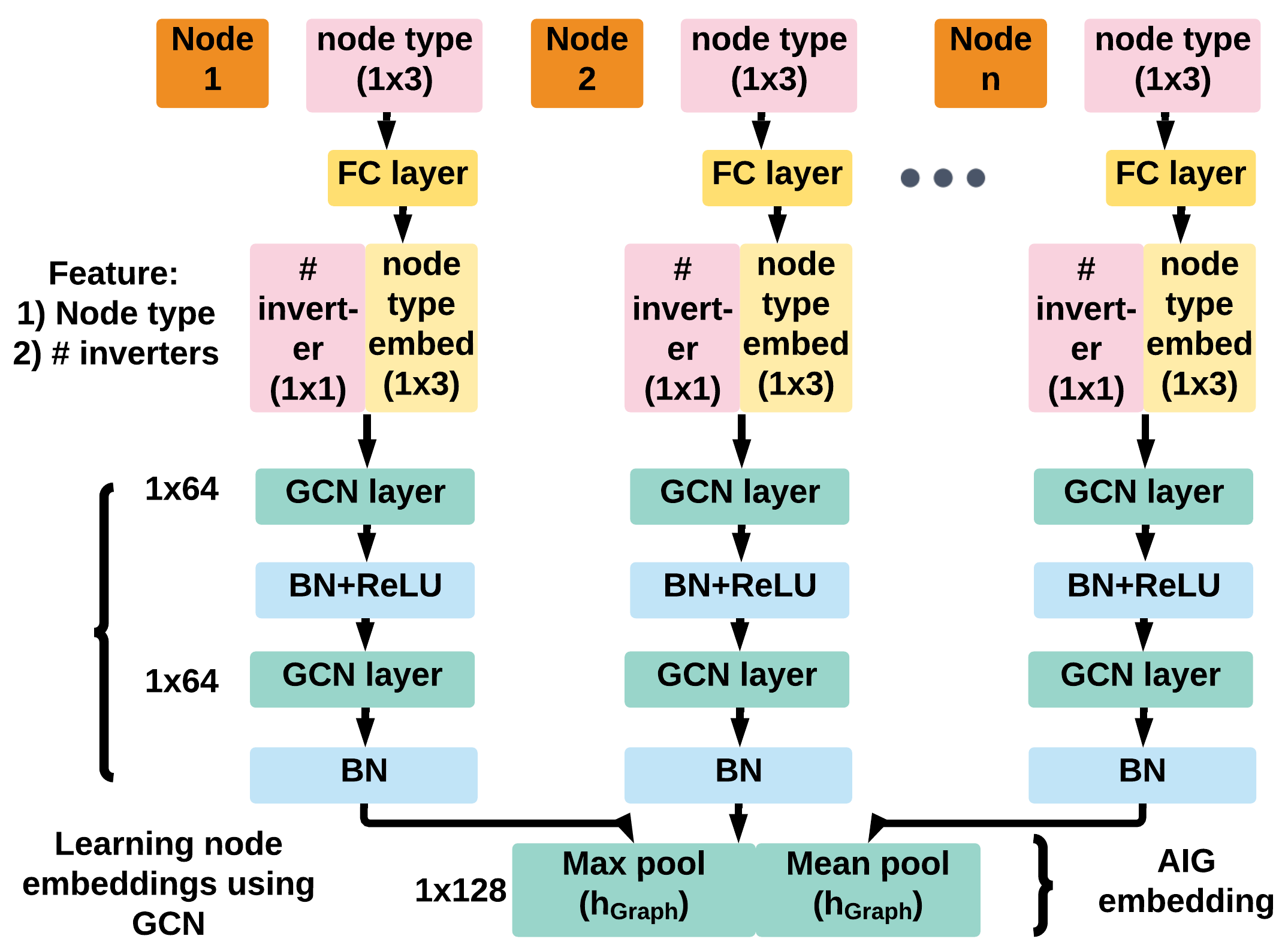}} \\
    \caption{Architecture for \ac{QoR} prediction. $GCN$: Graph convolutional layer. $FC$: fully connected layers. $BN$: Batch normalization layer.}
    \label{fig:OpenABCDnetwork}
\end{figure}

\subsubsection{AIG embedding}
We leverage \ac{GCN} to represent the original AIG graph. \ac{GCN} can learn and extract key features from \ac{DAG} structures; particularly subgraph structures that can be optimized via various synthesis transformations. \acp{GCN} make our model generalize over different \acp{AIG} structures. For \ac{AIG} embedding, we consider two features for each node: type and \# inverter in fan-in. The type of node can be primary input (PI), primary output (PO) or internal node. As shown in \autoref{fig:OpenABCDnetwork}a, we encode $node type$ and pass it via a linear layer. The initial node level features pass through two consecutive \ac{GCN} layers aggregating local neighbourhood information based on connectivity. We use batch-norm followed by ReLU operation after each \ac{GCN} layer. The graph level embedding is a readout like global pool operation on all node embeddings. We choose concatenation of average pooling and max pooling to produce the \ac{AIG} embedding $h_{AIG}$. Formally, the embeddings are defined as follows:
\begin{align}
    h^k_u &= \sigma ( W_k \times \sum_{u \cup N(u)} \frac{h^{k-1}_u}{\sqrt{N(u)}\times \sqrt{N(v)}} + b_{k} ) , k \in [1,2]\\
    h_{AIG} &= \frac{1}{|V|}\sum_{u \in V}h^k_u\parallel \max_{u} h^k_{u}
\end{align}
where $h^k_u$ is GCN embedding generated by $k$th layer of node $u$. $W_k$ and $b_k$ are GCN parameters and $\sigma$ is non-linear ReLU activation. $N(\cdot)$ denotes 1-hop neighbors.

\subsubsection{Synthesis recipe embedding}
In our work, we consider a fixed length synthesis recipe $\mathbf{L}$. We numerically encode the recipe using available synthesis transformation and pass it through a linear layer. Next, a set of one-dimensional \acp{CNN} with different kernel length extract features to produce ``synthesis recipe embedding'' ($h_{recipe}$). Mathematically, it can be shown as follows:

\begin{align}
    h_{recipe} &= concat_{i} (b_i + \sum_{l=j}^{j+M}s_{kl}W^l_{i}), i \in [1,4]
\end{align}
where $M$ is kernel length and ($W_i$,$b_i$) are filter parameters.

\begin{figure}[t]
    \centering
    \includegraphics[width=1.0\columnwidth]{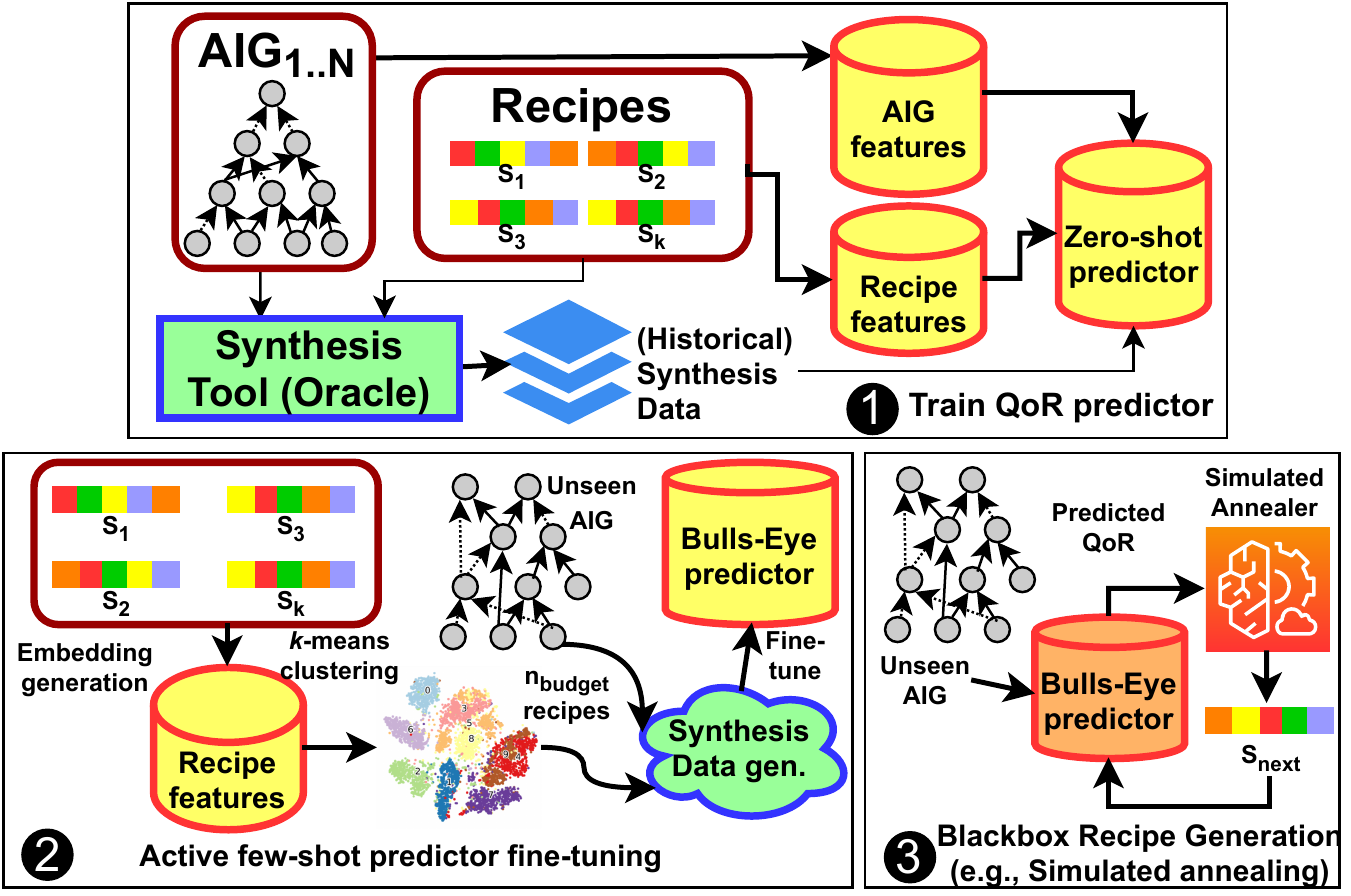}
    \caption{The \solution{} framework\label{fig:overall}}
\end{figure}

The \solution{} zero-shot predictor is created by concatenating  \ac{AIG} and synthesis recipe embeddings followed by four fully connected layers that output a \ac{QoR} prediction (~\autoref{fig:overall}). Assuming a training dataset, $\mathcal{D}_{train}$, comprising \acp{AIG} ($g_i$), synthesis recipes ($s_i$), and labels represented by number of nodes ($y_i$)---normalized to respective samples of design in the dataset---the graph-based regression learns a parameterized function $\hat{F}(\cdot;\theta_{ZS})$ by minimizing the loss function
\begin{align}
    \mathcal{L}(y,\hat{y}) = \frac{1}{|\mathcal{D}_{train}|}\sum_{(g_i,s_i) \in \mathcal{D}_{train}} \norm{y_i - \hat{F}(g_i,s_i;\theta_{ZS}) }_{2}^{2}.
\end{align}


\subsection{Fine-tuning \ac{QoR} predictor (\circletwo{}) }
One can use the zero-shot model to predict \ac{QoR} of different synthesis recipes on a \emph{new} \ac{AIG}. However, the accuracy of the model might be low if the features of the \emph{new} \ac{AIG} are different from prior data.
To address this problem, we propose to \emph{fine-tune} the zero-shot model using a \textbf{limited} number, $n_{budget}$, of synthesis runs on the new \ac{AIG}, where $n_{budget}$ is specified by the designer depending on the amount of time they are willing to spend. 


%
\textit{Fine-tuning} the zero-shot predictor avoids the need to train a model from for each new design, which was a critical roadblock for prior work. This strategy has been successful in a variety of other domains, for example, image classification~\cite{deng2009imagenet}, but has not been explored for ML-guided synthesis.
The next question we seek to answer is: \textit{How should the synthesis budget $n_{budget}$ be used?} 

We explore two solutions. The first solution, that we refer to as \textbf{FT+R} fine-tunes the zero-shot model using \emph{randomly} selected synthesis recipes. However, the aim of the designer is to pick best synthesis recipes giving wide variations in the synthesis output. This problem can be mapped similarly to an \textit{active learning} problem where label generation is a costly operation and therefore needs careful picking of data-points. Inspired by active learning approach~\cite{bodo2011active}, we propose a smarter solution, \textbf{FT+A}. Here, we \emph{cluster} the embeddings of the synthesis recipes used to train the zero-shot model (recall that each recipe has its own embedding denoted by $h_{recipe}$) into $n_{budget}$ clusters, and pick the cluster heads as our candidates. The intuition behind this idea is that the cluster heads represent maximally diverse and informative points in the solution space. 
We synthesize the unseen \ac{AIG} using the recipe cluster heads, generate labels and fine-tune $\hat{F}(\cdot;\theta_{ZS})$ to produce a high-quality \ac{QoR} predictor. 

\setlength{\textfloatsep}{-10pt}
\begin{algorithm}[t]
\caption{Active learning based model fine-tuning}
\label{algo:clusterProxy}
\KwData{Unseen AIG ($g_{U}$); Zero-shot predictor: $\hat{F}(\cdot;\theta_{ZS})$; synthesis recipes ($S$): $\{s_i, i \in [1,1500]\}$, Synthesis budget runs: $n_{budget}$} 
\KwResult{Active learning aided \ac{QoR} predictor: $\hat{F}(\cdot;\theta^{A}_{FT})$}
Create recipe embedding $h_{S} \leftarrow \{h_{s_i} \forall s_i \in S\}$ using $\hat{F}(\cdot,\theta_{ZS})$. \\
$S_{cluster} \leftarrow k$-means clustering using $n_{budget}$ heads on $h_{S}$.\\
Synthesize and generate labeled data $D_{U} \leftarrow \{(g_U,s_k,y_k)|s_k \in S_{cluster}, |S_{cluster}| = n_{budget}\}$.\\
Fine-tune learnable parameters $\theta_{FT}^A \leftarrow \theta_{ZS} - \alpha\nabla_{\theta}\mathcal{L}(\theta,\mathcal{D}_U)$

\end{algorithm} 
\setlength{\textfloatsep}{0pt}
\begin{algorithm}[t]
\caption{Simulated Annealing based recipe generator}
\label{algo:SAproxy}
\KwData{Initial temperature ($T_{initial}$); Max. oracle queries : $Q_{max}$; Fine-tuned \ac{QoR} predictor $\hat{F}(\cdot;\theta_{FT}^A)$, Design: ($AIG_{U}$)} 
\KwResult{Node optimized synthesis recipe: $s^* \in M^L$}
$T\gets T_{initial}$, $Q \leftarrow 0$, $s^*\gets$ \textbf{\Call{RandomRecipe}{$ $}} \\
\While{$Q < Q_{max}$}{
   $s_{next}\gets $ \textbf{\Call{NEIGHBOUR}{$T, s^*$}} \Comment{Neighbour recipes}\\
   $\Delta E\gets$ \textbf{\Call{ENERGY}{$s_{next},\hat{F}(\cdot;\theta_{FT}^A)$}} $-$ \textbf{\Call{ENERGY}{$s^*,\hat{F}(\cdot;\theta_{FT}^A)$}} \Comment{Energy $\downarrow \Rightarrow$ no. of nodes $\downarrow$}\\
   \If{$\Delta E < 0$ or \textbf{\Call{random}{$ $}} $<$ \textbf{\Call{ACCEPT}{$T,\Delta E$}}} {$s^*\gets s_{next}$}
   $T\gets$ \textbf{\Call{COOLING}{$T,s^*$}}; $Q \leftarrow Q+1$ \Comment{Annealing schedule}
}
	
\Return $s^*$ \Comment{Best synthesis recipe}
\end{algorithm}
\setlength{\textfloatsep}{\textfloatsepsave}


\subsection{Black-box Optimization (\circlethree{}) }
 
Although \solution{} (the zero-shot and fine-tuned predictors) can be used in a wide variety of black-box approaches (e.g., evolutionary algorithms), we focus on \ac{SA} adopted in \cite{simulatedAnnealing}. 
A good \ac{QoR} predictor provides fast and accurate feedback to the optimizer by estimating the \ac{QoR} of a synthesis recipe for an \ac{AIG}. 
As we will show in our experimental evaluation, \ac{QoR} predictor aided \ac{SA} can substantially explore the synthesis recipe space in a given time span. 
\autoref{algo:SAproxy} outlines the flow for generating synthesis recipes. 
The terminology in~\autoref{algo:SAproxy} is based on the \ac{SA} literature; readers can peruse \cite{simulatedAnnealing} for more background.

\section{Data generation and analysis}
\label{sec:dataGen}

\begin{table}[!tb]
\centering
\footnotesize
\setlength\tabcolsep{3pt}
\begin{tabular}{@{}lrrrr@{}}
\toprule
 & \multicolumn{4}{c}{Characteristics} \\ 
 \cmidrule(l){2-5} 
 \multirow{-2}{*}{Benchmark} & PI & PO & Nodes & Depth \\ \midrule
spi~\cite{opencores}  & 254 & 238 & 4219 & 35 \\
i2c\cite{opencores}  & 177 & 128 & 1169 & 15 \\
ss\_pcm\cite{opencores} & 104 & 90 & 462 & 10 \\
usb\_phy\cite{opencores} & 132 & 90 & 487 & 10 \\
sasc\cite{opencores} & 135 & 125 & 613 & 9\\
wb\_dma\cite{opencores} & {828} & {702} & {4587} & {29} \\
simple\_spi\cite{opencores} & {164} & {132} & {930} & {12} \\
pci\cite{opencores} &	{3429} &	{3157} &	{19547} &{29} \\
ac97\_ctrl\cite{opencores} &	{2339} &	{2137} &	{11464} & {11}\\
mem\_ctrl\cite{opencores} &	{1187} &	{962}& {16307} &{36} \\
des3\_area\cite{opencores} & {303} & {64} & {4971} & {30} \\
aes\cite{opencores} & {683} & {529} & {28925} & {27} \\
sha256\cite{mitll-cep} &	{1943} &	{1042} &	{15816} & {76} \\
{fir\cite{mitll-cep}} & {410} & {351} & {4558} & {47}\\
{iir\cite{mitll-cep}} & {494} & {441} & {6978} & {73} \\
{tv80\cite{opencores}} &	{636}	& {361} & {11328} &	{54} \\
{fpu\cite{balkind2016openpiton}} &	{632} &	{409} &	{29623} & {819}  \\
{dynamic\_node\cite{ajayi2019openroad}}  & {2708} & {2575} & {18094} & {33} \\
\midrule
apex1\cite{mcnc} & 45 & 45 &1577 & 14 \\ 
bc0\cite{mcnc} & 26 & 11 & 1592 & 31 \\
c1355\cite{iscas85} & 41 & 32 & 512 & 29 \\
c5315\cite{iscas85} & 178 & 123 & 1613 & 38 \\
c6288\cite{iscas85} & 32 & 32 & 2337 & 120 \\
c7552\cite{iscas85} & 207 & 107 & 2198 & 30 \\
dalu\cite{mcnc} & 75 & 16 & 1735 & 35 \\
i10\cite{abc} & 257 & 224 & 273 & 58 \\
k2\cite{mcnc} & 45 & 45 & 2289 & 22 \\
mainpla\cite{mcnc} & 27 & 54 & 5346 & 38 \\
div\cite{amaru2015epfl} & 128 & 128 &57247 & 4372 \\
log2\cite{amaru2015epfl} & 32 & 32 & 32060 & 444 \\
max\cite{amaru2015epfl} & 512 & 130 & 2865 & 287 \\
multiplier\cite{amaru2015epfl} & 128 & 128 & 27062 & 274 \\
sin\cite{amaru2015epfl} & 24 & 25 & 5416 & 225 \\
sqrt\cite{amaru2015epfl} & 128 & 64 & 24618 & 5058 \\
square\cite{amaru2015epfl} & 64 & 128 & 18484 & 250 \\
aes\_xcrypt\cite{aesxcrypt} &	{1975} &	{1805} &	{45840} & {43} \\	
{aes\_secworks\cite{aessecworks}}  &	{3087} &	{2604} &	{40778} & {42} \\
bp\_be\cite{bpsoc} &	{11592} & {8413} & {82514}	& {86} \\
wb\_conmax\cite{opencores} &	{2122} &	{2075} &	{47840} &{24}\\
ethernet\cite{opencores} &	{10731} & {10422} & {67164}& {34}\\
{jpeg\cite{opencores}} & {4962} & {4789} &	{114771} & {40}\\
{tiny\_rocket\cite{ajayi2019openroad}} &	{4561} &	{4181} &	{52315} & {80}\\
{picosoc\cite{balkind2016openpiton}} &	{11302} &	{10797} &	{82945} & 43 \\
vga\_lcd\cite{opencores} & {17322} & {17063} &	{105334} & {23} \\
\bottomrule
\end{tabular}
\caption{Open source design characteristics (unoptimized).  Primary Inputs (PI), Primary outputs (PO)\label{tab:structuralcharacteristics}}

\end{table}

\begin{figure}[!htb]
\centering
\includegraphics[width=\columnwidth]{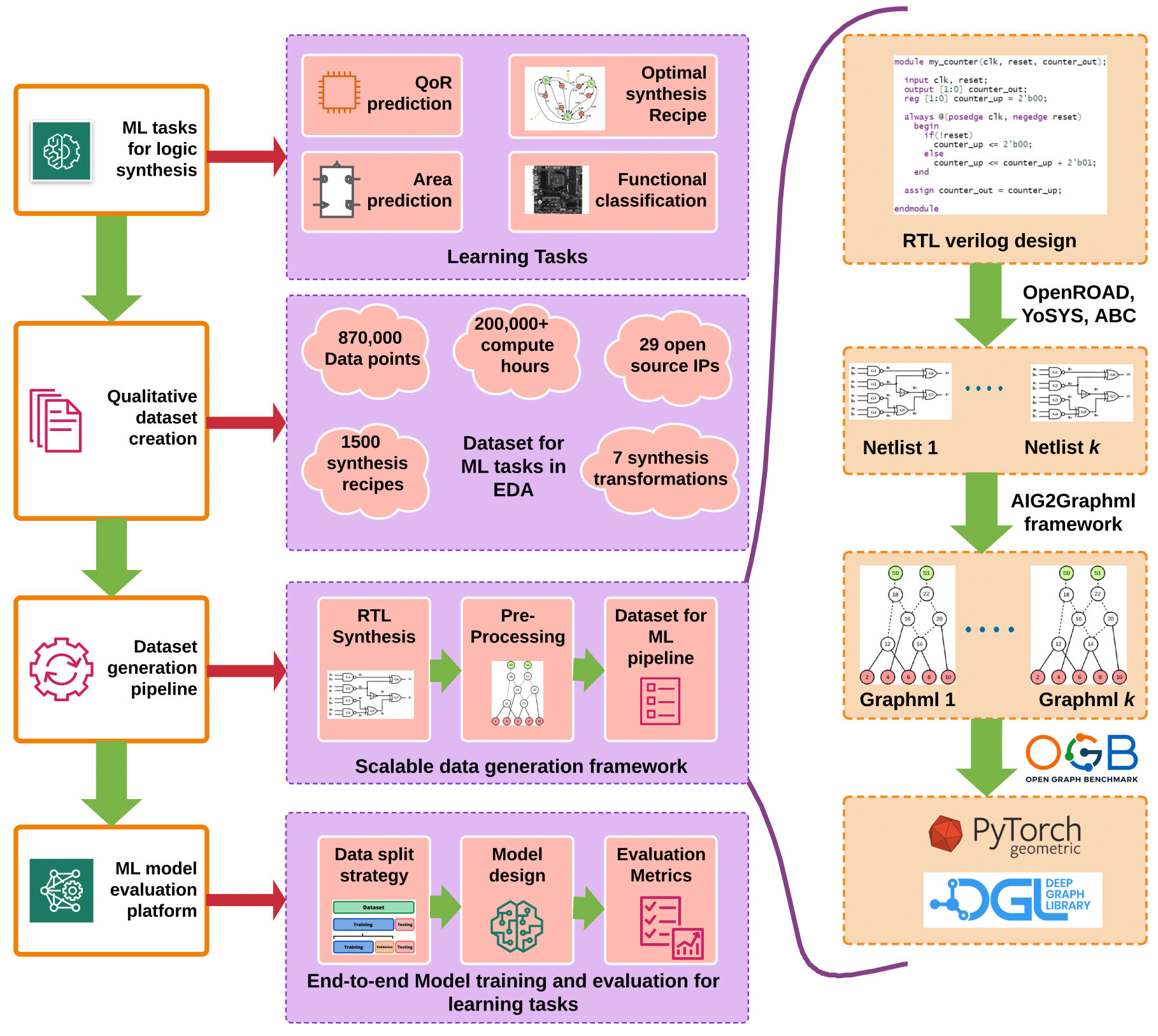}
\caption{OpenABC-D framework}
\label{fig:openabcdframework}
\end{figure}

For wider access and reproducible, we focus on open source EDA platforms as key to advancing \ac{ML} research in EDA.
Thus, we develop OpenABC-D framework(\autoref{fig:openabcdframework}): an end-to-end large-scale data generation framework for augmenting \ac{ML} research in circuit synthesis and making it freely available.
We applied the OpenABC-D framework on a set of open source designs to generate dataset for training our \solution{} predictor.

\begin{figure*}[!htb]
    \centering
        \hspace*{-0.2in}
    \subfloat[\label{fig:h1}Top $1\%$]{\includegraphics[width=0.66\columnwidth, valign=c]{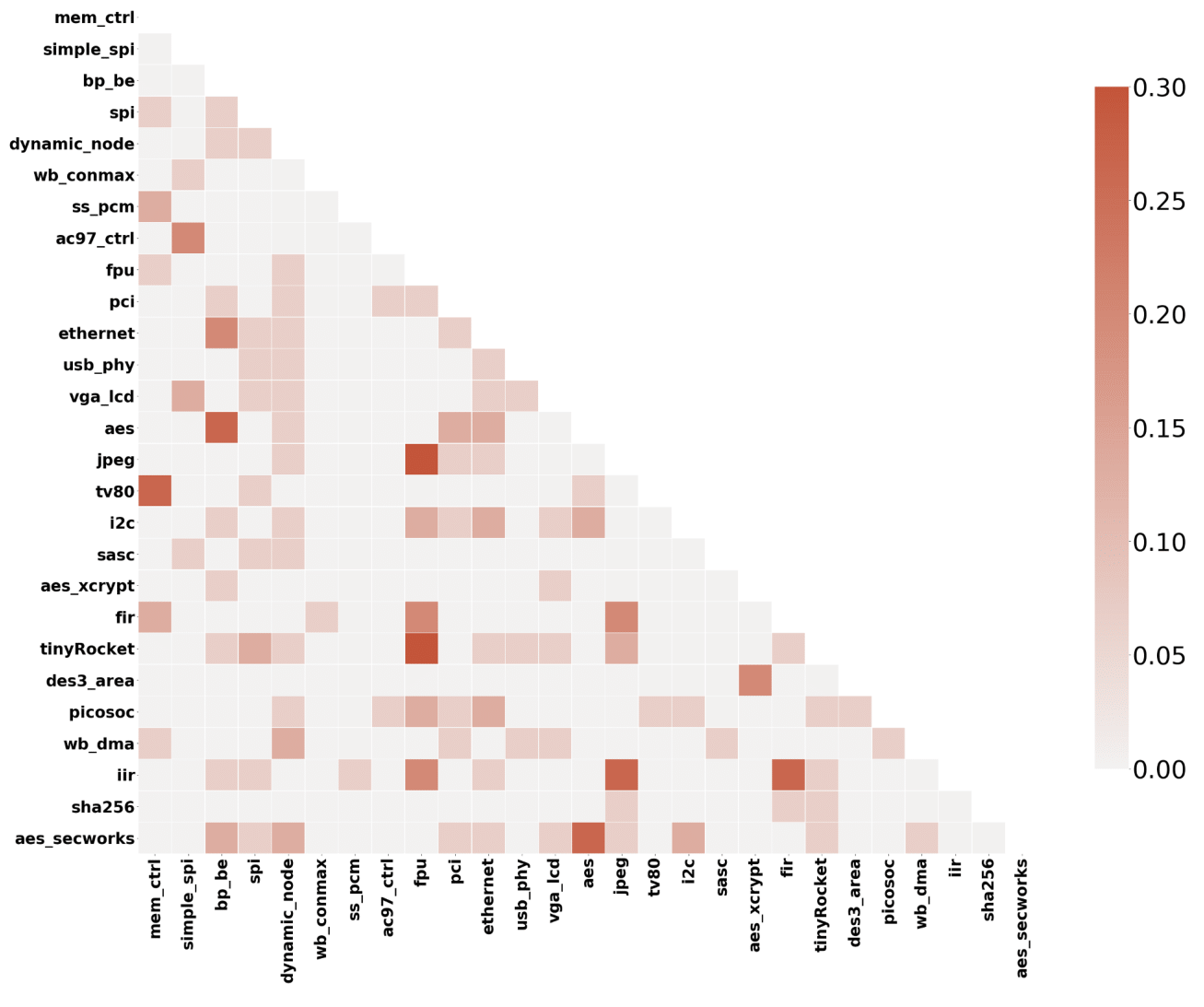}}
    \hspace*{-0.27in}
    \subfloat[\label{fig:h2}Top $5\%$]{\includegraphics[width=0.66\columnwidth, valign=c]{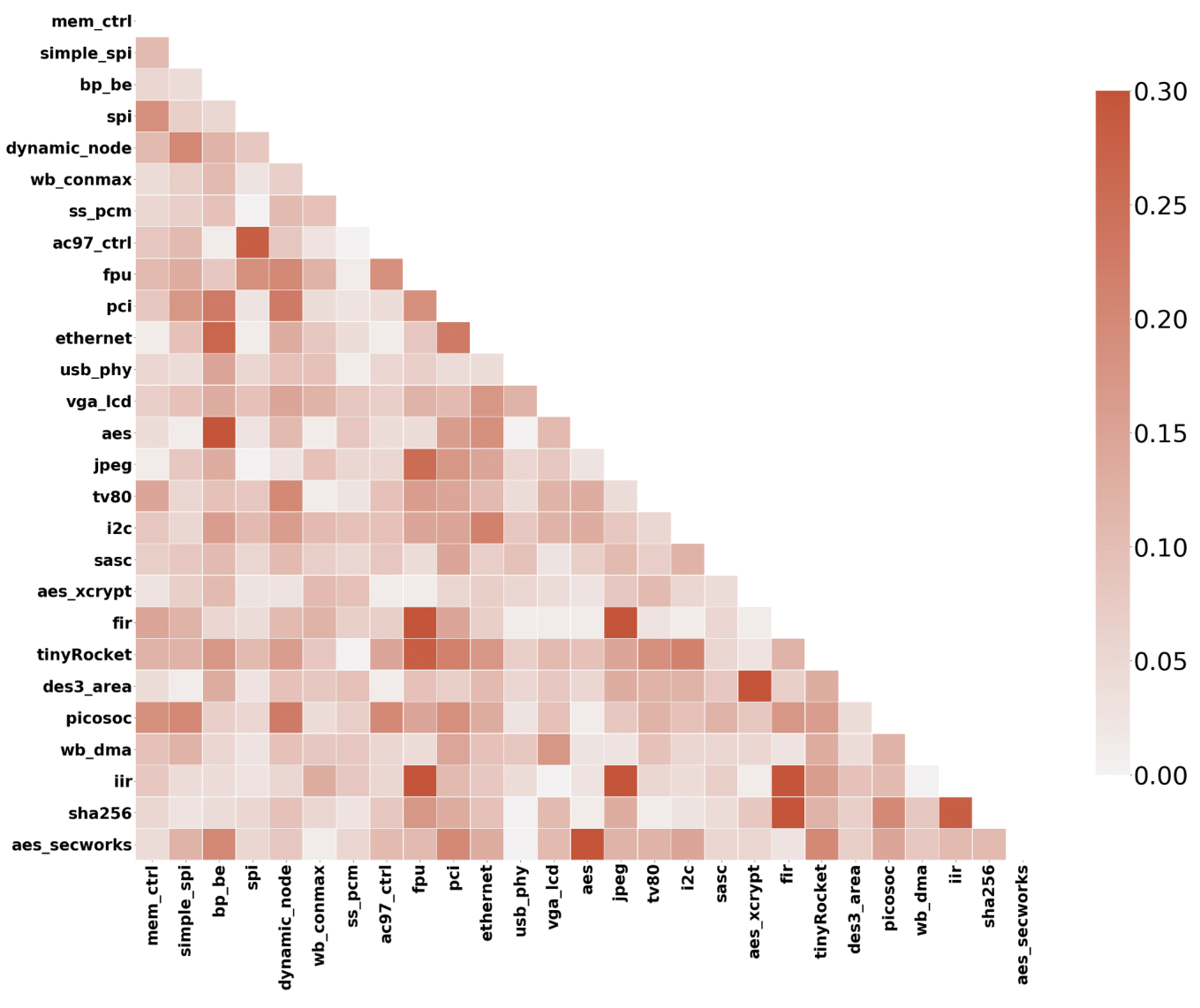}} 
       \hspace*{-0.27in}
    \subfloat[\label{fig:h3}Top $10\%$]{\includegraphics[width=0.66\columnwidth, valign=c]{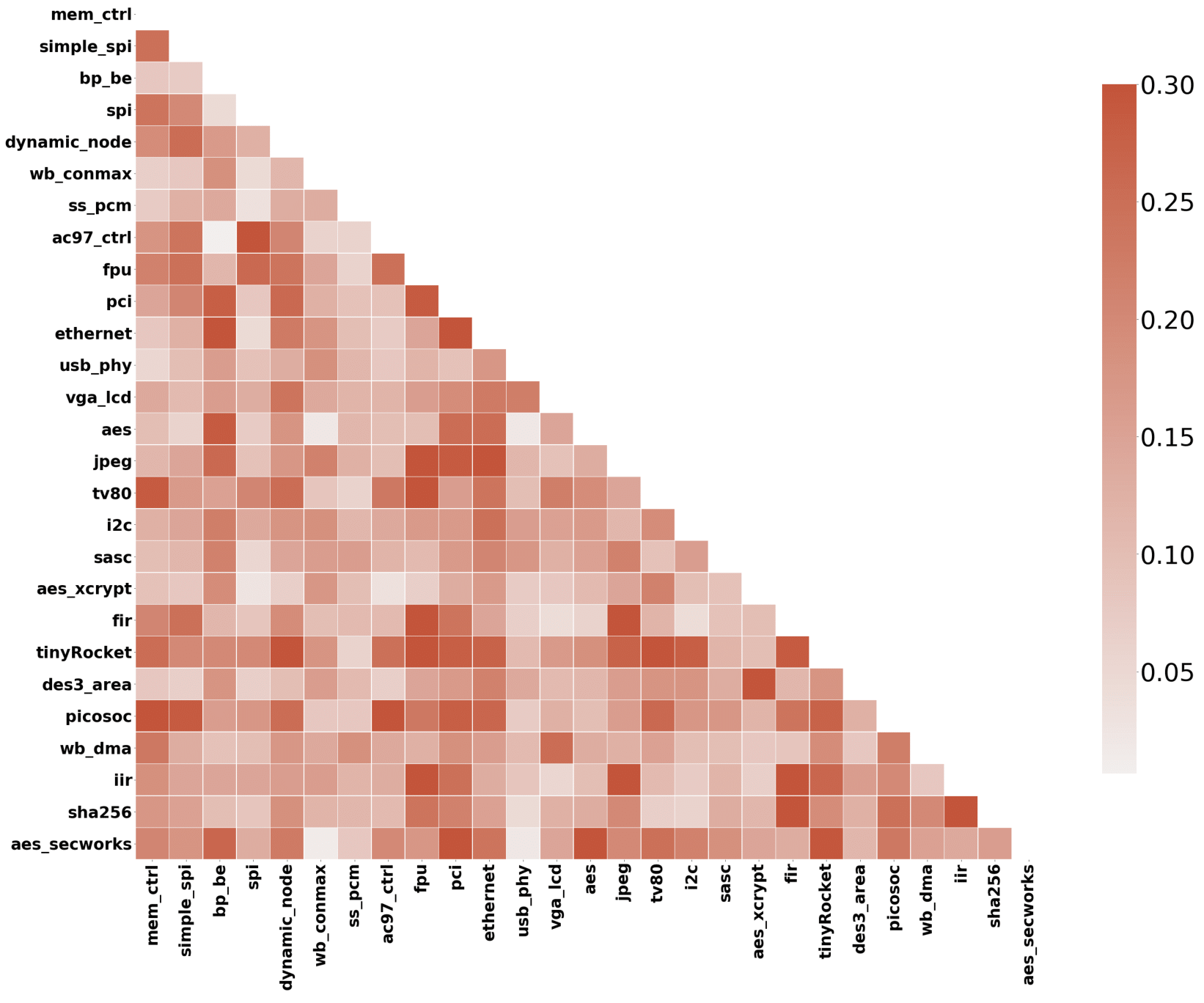}}
  \caption{Correlation plots amongst top $k\%$ synthesis recipes various IPs. Darker colors indicate higher similarity between the top synthesis recipes for the pair of IPs.}
    \label{fig:synthesis-tops}
\end{figure*} 

\subsection{Data generation}

To produce the OpenABC-dataset, we use 44 open source designs with a wide range of functions. 
In the absence of a preexisting dataset like ImageNet~\cite{deng2009imagenet} that represents many classes of designs, we hand-curated designs of different functionalities from MIT LL labs CEP~\cite{mitll-cep}, OpenCores~\cite{opencores}, IWLS~\cite{albrecht2005iwls}, ISCAS~\cite{iscas85,iscas89}, MCNC~\cite{mcnc} and EPFL benchmarks~\cite{amaru2015epfl}. These benchmarks include more complex and functionally diverse compared to only ISCAS~\cite{iscas85,iscas89} and EPFL~\cite{amaru2015epfl} benchmarks that are used in prior work and represent large, industrial-sized designs. In the context of EDA, functionally diverse IPs should have diverse AIG structures (e.g., tree-like, balanced, and skewed) that mimic the distribution of real hardware designs. \autoref{tab:structuralcharacteristics} summarizes the structural characteristics of the data before synthesis (without optimizations). These designs are functionally diverse -- bus communication protocols, computing processors, digital signal processing cores, cryptographic accelerators and system controllers. We prepared $K = 1500$ synthesis recipes each $L = 20$ long, comprising a mix of \texttt{rewrite, re-substitute, refactor, rewrite -z, resub -z, refactor -z} and \texttt{balance} transformations, consistent with prior works~\cite{mlcad_abc,iccad_2021,drills}. The synthesis recipes were prepared by randomly sampling from the set of synthesis transformations, assuming a uniform distribution. 
We ran synthesis with \textit{abc} using server-grade Intel processors for $200,000+$ computation hours to generate the labeled data. We analyzed the top $k$ synthesis recipes (in terms of node optimization) by varying $k=15$, $75$ and $150$ ($1\%$ to $10\%$). We found that the commonality of the top performing recipes across any two designs is less than 30\% (\autoref{fig:synthesis-tops}). This indicate that the \ac{AIG} structures of designs are diverse and  synthesis recipes perform differently across them.

\subsection{Preprocessing for \ac{ML}}
\label{subsec:datagen-mlpipeline}

This stage involves preparing the circuit data for use with any \ac{ML} framework. 
We use pytorch-geometric APIs to create data samples using \ac{AIG} graphs, synthesis recipes, and number of nodes and delay of synthesized \acp{AIG}. The dataset is available using a customized dataloader for easy handling for preprocessing, labeling, and transformation. We create a script that helps partition the dataset (e.g., into train/test) based on user specified learning tasks. 

\section{Experimental Setup and Evaluation}
\label{sec:exp}

As we discussed in~\autoref{sec:intro}, \solution{} illustrates ideas of fine-tuning, active learning, and scalability for ML-guided logic synthesis. We focus our experimental evaluation on characterizing \solution{} in these dimensions. 

\subsection{Setup}
We evaluate the efficacy of the \solution{} predictor and the quality of the recipes found using \solution{} on OpenABC-dataset. We emphasize that our work scales to larger designs compared to prior work.

%

For assessing quality of a recipe, we use the following metrics for comparison. These metrics provide a qualitative indicator on number of nodes optimized in \ac{AIG} and relative improvement compared to baselines:
\begin{itemize}
    \item Reduction factor (\%): This metric depicts a ratio of number of nodes optimized post synthesis to number of nodes in original netlist. This is denoted by:
    \begin{align}
     rf (\%) &= \frac{N_{original}-N_{synthesized}}{N_{original}}*100 \%
    \end{align}
    \item Relative Improvement (\%): This metric measures relative performance of reduction factor of a proposed technique compared to baseline technique. This is denoted by:
    \begin{align}
     RI (\%) &= \frac{rf_{proposed}-rf_{baseline}}{rf_{baseline}}*100 \%
    \end{align}
\end{itemize}

We use these metrics to compare the synthesis output quality using recipes generated by simulated annealing with \solution{} QoR predictor. As an intermediate result, we also evaluate the quality of the \solution{} \ac{QoR} predictors. For this, we use a reference set of 1500 recipes and corresponding synthesis results  as the ground-truth \ac{QoR}. We draw from this reference set the various train/test recipe splits for evaluation.

In our experiment, we train the zero-shot predictor (\autoref{tab:network-architecture-params}) on $18$ \emph{small} designs ($\leq 30k$ nodes, upper half of ~\autoref{tab:structuralcharacteristics}), and fine-tune and evaluate it on $26$ designs ($10$ large designs with $41$k$-114$k nodes) from the OpenABC dataset. 
We evaluate the predictor on designs from prior works for comparisons to the \ac{SOTA}. 

To train our models, we use \ac{MSE} as the loss function and normalized number of nodes as labels (normalized across each benchmark). 
The hyperparameters used include Adam optimizer, learning rate of $0.01$ and batch size of $64$.
During fine-tuning, we set the learning rate to $0.01$ for fully connected layers and $0.001$ for \ac{GCN} and 1-D convolution layers. We perform experiments on a 2.9GHz Intel Xeon CPU with 384GB RAM and NVIDIA V100 GPU. We train the zero-shot model (\circleone{}) for 80 epochs with a runtime of about 12 hours as a \textit{one-time} effort. This relatively low one-time effort is because we only use small netlists to train the zero-shot predictor; surprisingly, we find it transfers well to new and unseen large netlists. 
Once the model zero-shot model trained, it can be directly used or fine-tuned for \emph{any} new design. 
We perform fine-tuning for 20 epochs that takes  less than $200$ seconds for each new design that we evaluate, including the large designs. 

\begin{table}[t]
\centering
\caption{GCN architecture. I: Input dimension, L1, L2: dimension of GCN layers: f: \# filters, k: kernels, s: stride, \# l: \#FC layers\label{tab:network-architecture-params}}
\resizebox{\columnwidth}{!}{%
\begin{tabular}{@{}cccccccccc@{}}
\toprule
\multicolumn{4}{c}{AIG Embedding} & \multicolumn{4}{c}{Recipe Encoding} & \multicolumn{2}{c}{FC Layers} \\ \cmidrule(lr){1-4}\cmidrule(lr){5-8}\cmidrule(lr){9-10} 
I & L1 & L2 & Pool & I & \#f & k & s & \# l & arch\\ \midrule
4 & 64 & 64 & Max+Mean & 60 & 4 & 12,15,18,21 & 3 & 4 & 190-512-512-512-1\\
\bottomrule
\end{tabular}%
}
\end{table}

\subsection{Evaluating \ac{QoR} predictor}
\subsubsection{Experiment design} 
We first examine how well the predictors enables accurate \ac{QoR} prediction by comparing different variants of the \solution{} predictor. 
We evaluate and compare their predictions on the reference set of randomly selected synthesis recipes (for which we have actual synthesis results). We capture this as the predictor's \textit{Commonality factor}, which is defined as $\frac{n(P \cap A)}{n(P)}$, where $A$ denotes top k\% performing synthesis recipes (actual) and $P$ denotes top k\% synthesis recipes (predicted). 
%
%
As mentioned earlier, we use $9$ large benchmarks from OpenABC-D framework and $17$ benchmarks from \ac{SOTA} works for evaluation. 
The models are: 

\begin{itemize}
    \item \textit{Standalone} \ac{QoR} predictor is trained from scratch using $n_{budget}$ samples from synthesizing the target design. 
    \item \textit{Zero-shot} ($ZS$) \ac{QoR} predictor is trained only on small designs from the OpenABC-D synthesis dataset.
    \item \textit{Fine-tuning + random} ($FT+R$) \ac{QoR} predictor, which is fine-tuned on randomly chosen $n_{budget}$ recipes and corresponding synthesis results as fine-tuning (calibration) data. 
    \item \textit{Fine-tuning + active} 
    ($FT+A$) \ac{QoR} predictor, which is prepared 
    by fine-tuning on synthesis data from $n_{budget}$ most-informative recipes (generated from k-means clustering of the $1500$ reference set recipe embeddings).
\end{itemize}


\begin{figure}[!ht]
    \centering
    \subfloat[][Standalone]{\includegraphics[width=\columnwidth]{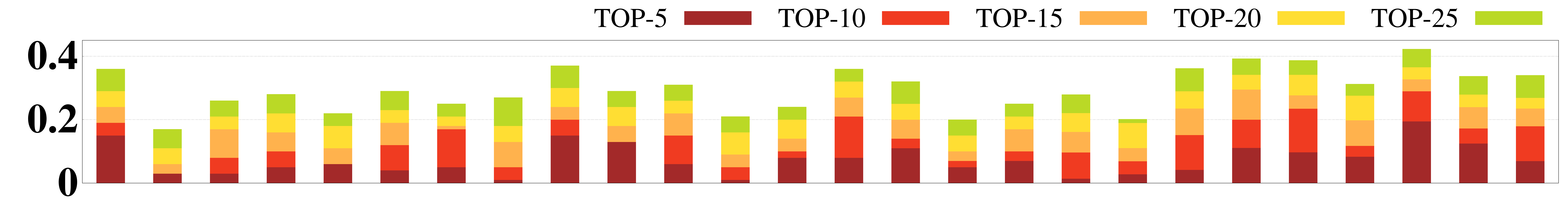}} \\
    \subfloat[][\solution{} (Zero-shot)]{\includegraphics[width=\columnwidth]{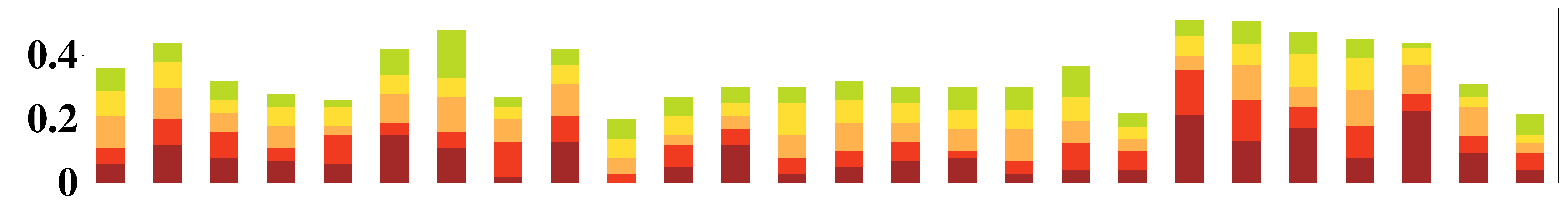}} \\
    \subfloat[][\solution{} (Fine-tune + Random)]{\includegraphics[width=\columnwidth]{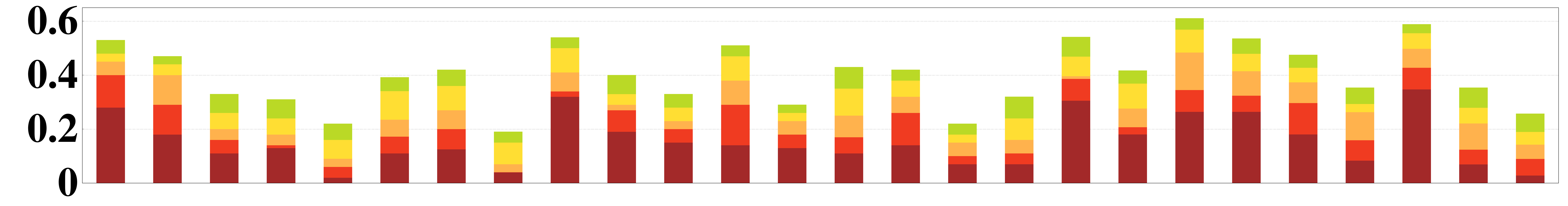}} \\
    \subfloat[][\solution{} (Fine-tune + Active)]{\includegraphics[width=\columnwidth]{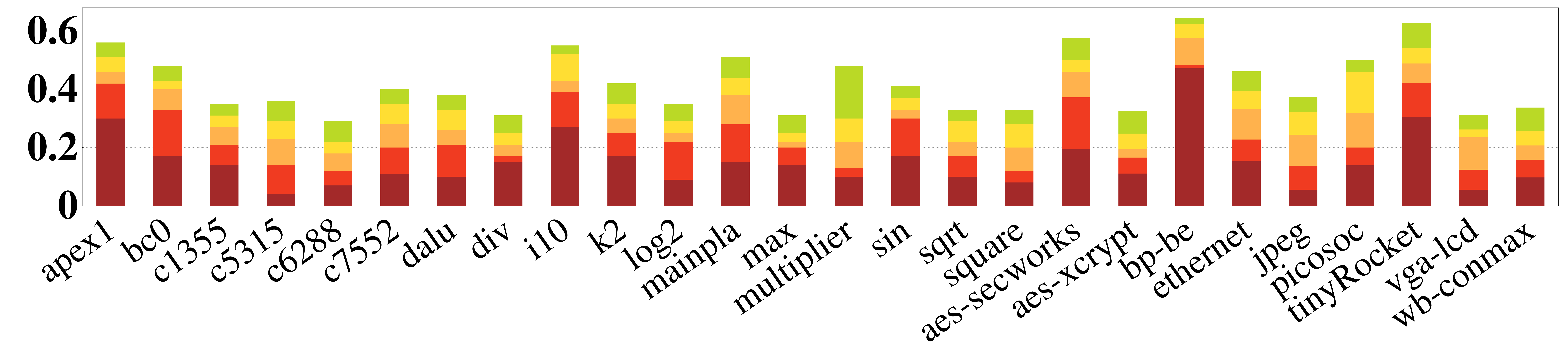}}
    \caption{Commonality factor (y-axes) among top k\% recipes between actual synthesis output and predictions for  designs (x-axes). ($\uparrow$ is better)}
    \label{fig:commonalityFactor}
\end{figure}

\subsubsection{Analysis of Results} 
\autoref{fig:commonalityFactor} presents the performance of all \ac{QoR} predictors using commonality factor as the metric across various top-$k$\% values.
For example, \autoref{fig:commonalityFactor}(d) shows that \solution{} using fine-tuning and active learning achieves 0.46 for top-5\% on the  \textit{bp-be} design. This indicates 46\% of top 5\% performing synthesis recipes (out of 1450) on \textit{bp-be} is common with top 5\% synthesis recipes predicted by \ac{QoR} predictor. 
Generally, we observe fine-tuned \ac{QoR} predictors yield better results (better commonality factor for every top k\%) compared to standalone and zero-shot predictors. 
A high commonality factor denotes that the model can infer better \ac{QoR} achieving recipes successfully; pre-trained predictors are indeed beneficial as the basis for fine-tuning \ac{QoR} predictor. 


\subsection{\ac{QoR} of \solution{} generated recipes}
\subsubsection{Experiment design} A good \ac{QoR} predictor accurately predicts the quality of a synthesis recipe for a given design to provide good feedback in an optimization scheme. 
\solution{} uses \ac{QoR} predictors in lieu of actual synthesis runs as the backend engine (``evaluator'') of an \ac{SA}-based recipe generator.
For a fair evaluation, we set two baselines to compare the quality of recipes generated using a \ac{QoR} predictor: 1) $k$-repeat application of \ac{SA}-generated recipe using actual synthesis runs as the evaluator 2) resyn2$^*$: repeated application of resyn2 synthesis recipe. In our experiments, we set resyn2 as starting seed for simulated annealing and $k=25$. For a fair performance evaluation, we set a threshold of $1000$ calls to evaluator of simulated annealing and compare runtime speed up of \ac{QoR} predictor versus conventional actual synthesis. We compare \solution{} on two definitive aspects: 1) Run-time speedup of simulated annealing using \solution{} predictor instead of actual synthesis, 2) Relative performance improvement of \solution{} compared to resyn2$^*$ for same runtime. We set the initial temperature $T_{start}=7250$ and distorted \textit{Caunchy-Lorentz} distribution as visiting distribution as part of simulated annealing.

\begin{table*}[t]
\centering
\caption{Reduction factor (\%) obtained by our work and prior works}
\footnotesize
\label{tab:bullseye_sota}
\setlength\tabcolsep{3pt}
\resizebox{2\columnwidth}{!}{%
\begin{tabular}{lrrrrrrrrrrrrrrrrrrr}
\toprule
& & \multicolumn{8}{c}{Reduction factor (\%)} & \multicolumn{5}{c}{RI w.r.t. resyn2$^*$ (\%)} & \multicolumn{5}{c}{RI w.r.t. $SA_{1k}$ (\%)} \\ \cmidrule(lr){3-10} \cmidrule(lr){11-15} \cmidrule(lr){16-20}
& & \multicolumn{3}{c}{Baseline} & \multicolumn{2}{c}{SOTA} & \multicolumn{3}{c}{\solution} & \multicolumn{2}{c}{SOTA} & \multicolumn{3}{c}{\solution} & \multicolumn{2}{c}{SOTA} & \multicolumn{3}{c}{\solution} \\
\cmidrule(lr){3-5} \cmidrule(lr){6-7} \cmidrule(lr){8-10} \cmidrule(lr){11-12} \cmidrule(lr){13-15} \cmidrule(lr){16-17} \cmidrule(lr){18-20}
\multirow{-3}{*}{Design} & \multirow{-3}{*}{Nodes} & \begin{tabular}[c]{@{}c@{}}{resyn2}\\ {(x2)}\end{tabular} & \begin{tabular}[c]{@{}c@{}}{resyn2$^*$}\end{tabular} & $SA_{1k}$ & \cite{mlcad_abc} & \cite{iccad_2021} & ZS & FT+R & FT+A  & \cite{mlcad_abc} & \cite{iccad_2021} & ZS & FT+R & FT+A & \cite{mlcad_abc} & \cite{iccad_2021} & ZS & FT+R & FT+A\\
\midrule
apex1 &  1577 & 26.44 & 29.93 & 35.89 & 28.66 & 28.97 & 30.37 & 37.09 & 40.39 & -4.23 & -3.17 & 1.48 & \btbf{23.94} & \rtbf{34.95} & -20.14 & -19.25 & -15.37 & 3.35 & 12.54\\
bc0 &  1592 & 44.40 & 45.03 & 53.14 & 48.49 & 48.05 & 45.97 & 48.86 & 53.39 & 7.67 & 6.69 & 2.09 & \btbf{8.51} & \rtbf{18.54} & -8.74 & -9.57 & -13.47 & -8.03 & 0.47\\
c1355 &  512 & 23.04 & 23.24 & 23.42 & 23.42 & 23.43 & 23.43 & 23.43 & 23.82 & 0.84 & \btbf{1.69} & \btbf{1.69} & \btbf{1.69} & \rtbf{3.38} & 0 & 0.84 & 0.84 & 0.84 & 2.52  \\
c5315 & 1613 & 20.08 & 20.58 & 22.56 & 20.45 & 26.53 & 22.50 & 22.56 & 22.75 & -0.60 & \rtbf{28.91} & 9.33 & 9.63 & \btbf{10.54} & -9.34 & 17.58 & -0.27 & 0 & 0.82 \\
c6288 & 2337 & 19.98 &	19.98	& 20.03 &	19.98&	19.98 &	19.98 &	19.98&	20.03&	0&	0&	0&	0&	\rtbf{0.25}&	-0.25&	-0.25&	-0.25&	-0.25&	0\\
c7552 & 2198 & 35.53&	36.08&	38.58&	36.26&	- &	37.31&	38.35&	38.81&	0.5&	- &	3.41&	\btbf{6.29} &	\rtbf{7.57}&	-6.01&	- &	-3.29&	-0.6&	0.6\\
dalu & 1735 & 39.37&	39.6&	46.51&	42.31&	40.58&	42.36&	45.24&	47.72&	6.84&	2.47&	6.97&	\btbf{14.24} &	\rtbf{20.51} &	-9.03&	-12.75&	-8.92&	-2.73&	2.6 \\
i10 & 2273  & 25.43	&25.43&	27.54&	27.63&	- &	26&	26.97&	27.76&	8.65&	- &	2.24&	\btbf{6.06} &	\rtbf{9.16} &	0.33&	- &	-5.59&	-2.07&	0.8\\
k2 & 2289 & 44.82&	46.48&	48.06&	47.53&	47.66&	47.79&	50.98&	52.08&	2.26&	2.54&	2.82&	\btbf{9.68} &	\rtbf{12.05} &	-1.1&	-0.83&	-0.56&	6.08&	8.36\\
mainpla & 5346 & 34.12&	34.46&	40.7&	35.69&	35.3&	37.06&	40.12&	41.75&	3.57&	2.44&	7.54&	\btbf{16.42}	& \rtbf{21.15}	&-12.31	&-13.27	& -8.94	-& 1.43	& 2.58\\
\midrule
\specialcell{Time (in s)} & - & 5.338 & 3932.8 & 3932.8 & $14946.4^{\dagger}$ & $1387.9^{\dagger}$ & 529.6 & 1250.53 & 1333.91 \\
\bottomrule
\end{tabular}
}\\
$\dagger$ \scriptsize 
Runtime scaled on our machine based on data provided by authors~\cite{iccad_2021} for synthesis runs during training. Total training runtime will be higher than reported here.
\end{table*}

\begin{table}[!htb]
\centering
\caption{Reduction factor (\%) obtained by our work on large benchmarks}
\footnotesize
\label{tab:bullseye_large}
\setlength\tabcolsep{2.5pt}
\resizebox{\columnwidth}{!}{%
\begin{tabular}{lrrrrrrrrrrr}
\toprule
& & \multicolumn{6}{c}{Reduction factor (\%)} & \multicolumn{3}{c}{RI w.r.t. resyn2$^*$} \\ \cmidrule(lr){3-8} \cmidrule(lr){9-11}
& & \multicolumn{3}{c}{Baseline} & \multicolumn{3}{c}{\solution} & \multicolumn{3}{c}{\solution} \\
\cmidrule(lr){3-5} \cmidrule(lr){6-8} \cmidrule(lr){9-11}
\multirow{-3}{*}{Design} & \multirow{-3}{*}{Nodes} & \begin{tabular}[c]{@{}c@{}}{resyn2}\\ {(x2)}\end{tabular} &\begin{tabular}[c]{@{}c@{}}{resyn2$^*$}\end{tabular} & $SA_{1k}$ & ZS & FT+R & FT+A  & ZS & FT+R & FT+A \\
\midrule
div & 57247 & 28.88	& 28.92 & 64.11 &	60.42 &	62.06 &	64.19 &	108.92 &	114.59 &	121.96 \\
log2 & 32060 & 8.74 &	8.80&	9.39&	8.82&	9.29&	9.42&	0.23&	5.57&	7.05 \\
max & 2865 & 1.19&	1.19&	1.50	&1.19&	1.29&	1.29&	0.00&	8.40	&8.40 \\
multiplier & 27062 & 9.27&	9.27&	10.14&	9.97&	10.14&	10.14&	7.55&	9.39&	9.39 \\
sin & 5416 & 7.42&	7.5	&7.75&	7.85&	8.23&	7.98&	4.67&	9.73&	6.40 \\
sqrt & 24618 & 21.05&	21.05&	22.42&	21.64&	22.07&	22.43&	2.80&	4.85&	6.56 \\
square & 18484 & 10.37&	10.37&	14.50&	12.89&	14.36&	14.57&	24.30&	38.48&	40.50 \\
aes\_secworks & 40778 & 23.5&	27.23&	27.63&	27.63&	27.9&	27.93&	1.47&	2.46&	2.57 \\
aes\_xcrypt & 45850 & 9.01&	16.03&	16.19&	16.18&	16.20&	16.35&	0.94&	1.06&	2.00\\
bp\_be & 82514 & 4.92&	5.05&	5.16&	5.34&	5.68&	5.96&	5.74&	12.48&	18.02\\
ethernet & 67164 & 3.30&	3.44&	3.47&	3.46&	3.47&	3.48&	0.58&	0.87&	1.16\\
jpeg & 114771 & 9.18&	9.37&	11.15&	11.13&	11.21&	11.32&	18.78&	19.64&	20.81\\
picosoc & 82945 & 7.40&	7.54&	7.74&	7.68&	7.93&	8.04&	1.86&	5.17&	6.63 \\
tinyRocket & 52315 & 21.30&	21.80&	22.26&	22.03&	22.22&	22.4&	1.06&	1.93&	2.75 \\
vga\_lcd & 105334 & 1.62&	1.64&	1.71&	1.67&	1.72&	1.73&	1.83&	4.88&	5.49 \\
wb\_conmax & 47840 & 6.72&	6.89&	9.35&	9.26&	10.05&	10.44&	34.40&	45.86&	51.52\\
\bottomrule
\end{tabular}%
}
\end{table}

\begin{figure}[!ht]
    \centering
    \subfloat[][Relative performance with resyn2$^*$]{\includegraphics[width=\columnwidth]{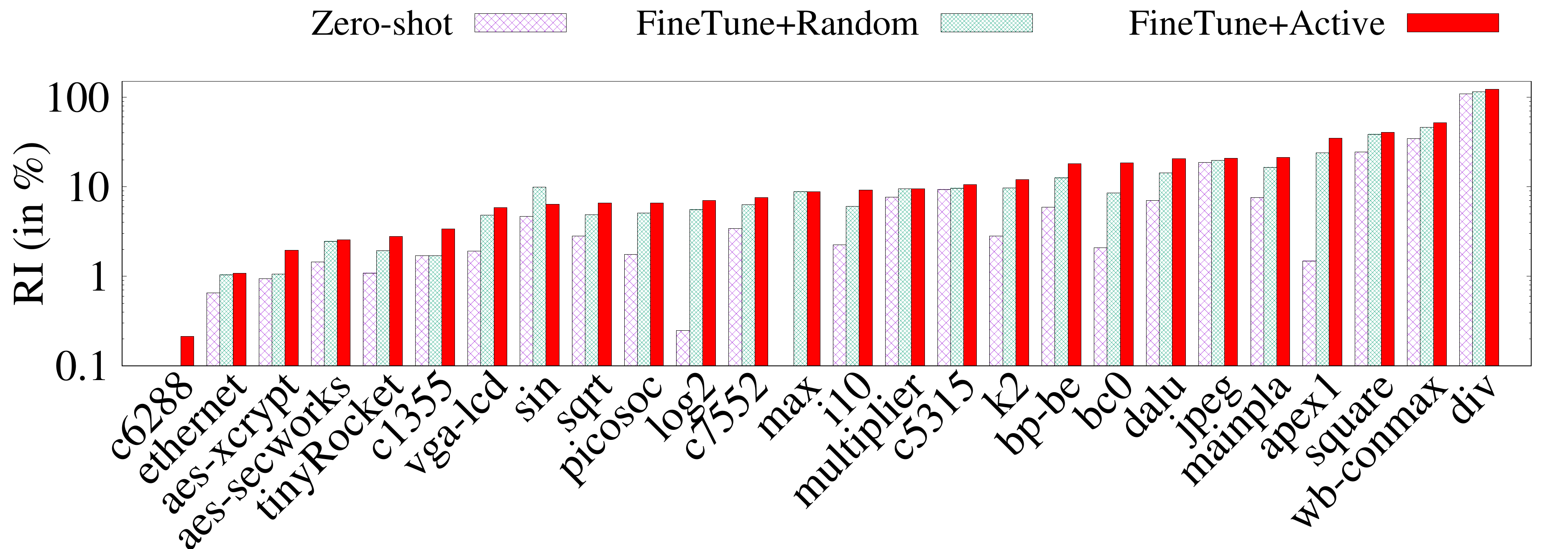}} \\
    \subfloat[][RI-speedup compared to conventional SA]{\includegraphics[width=\columnwidth]{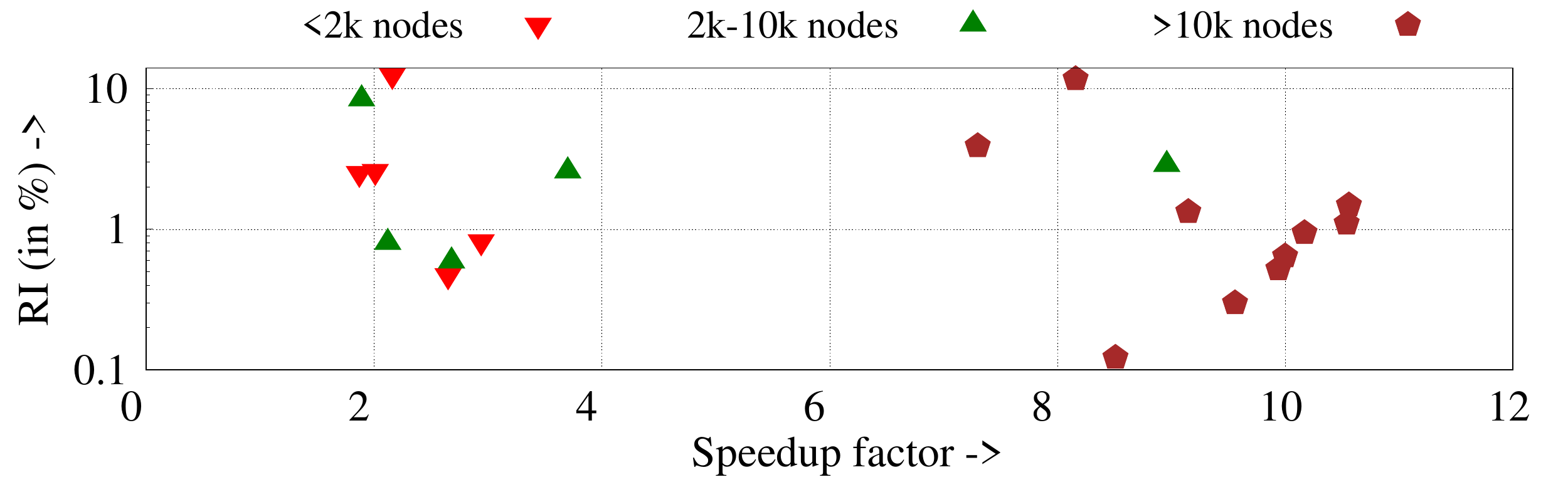}}
    \caption{Bullseye performance compared to resyn2$^*$ and conventional simulated annealing}
    \label{fig:bullseyeQualitative}
\end{figure}

\begin{table*}[!ht]
\centering
\caption{Timing complexity of baseline and \solution{} framework. \textit{FT}: Time consumed for finetuning the model, \textit{SA}: Time taken for simulated annealing, and \textit{synth}: Time taken for $k=25$ repeated application of synthesis recipe}
\footnotesize
\label{tab:bullseye_time}
\setlength\tabcolsep{3.5pt}
\resizebox{2\columnwidth}{!}{%
\begin{tabular}{lrrrrrrrrrrrrrrrrrr}
\toprule
& \multicolumn{15}{c}{Time taken (s)} & \multicolumn{3}{c}{Speedup} \\ 
\cmidrule(lr){2-16} \cmidrule(lr){17-19}
&  & \multicolumn{3}{c}{Baseline ($SA_{1k}$)} & \multicolumn{3}{c}{ZS} & \multicolumn{4}{c}{FT+R} & \multicolumn{4}{c}{FT+A} \\
\cmidrule(lr){3-5} \cmidrule(lr){6-8} \cmidrule(lr){9-12} \cmidrule(lr){13-16} 

\multirow{-3}{*}{Design} & \multirow{-2}{*}{\begin{tabular}[c]{@{}c@{}}{resyn2}\\ {(x2)}\end{tabular}} & SA & synth & total & SA & synth & total & FT & SA & synth & total & FT & SA & synth & total & \multirow{-2}{*}{ZS} & \multirow{-2}{*}{FT+R} & \multirow{-2}{*}{FT+A} \\
\midrule
apex1&	0.4 &	216.5&	13.2 &	229.7 &	35.5 &	11.3 &	46.8 &	48.3 &	30.2&	13.5	&92.0 &	53.1 &	37.6&	15.9 &	106.6	& 5.0 &	2.5&	2.2 \\
bc0	&0.3	&235.2 &	11.2&	246.4	&36.9	&9.8 &	46.7 &	42.2 &	30.2&	12.3&	84.7 &	46.5 &	37.5&	9.1 &	93.0 &	5.3 &	2.9 &	2.7 \\
c1355&	0.2 &	135.3&	5.4 &	140.7 &	37.5&	5.5 &	43.0 &	40.1 &	28.0 &	4.9 &	73.0 &	44.1 &	26.1 &	5.2&	75.4 &	3.3&	1.9&	1.9 \\
c5315&	0.5&	287.1&	16.5 &	303.6 &	32.6 &	17.5 &	50.1 &	49.6 &	33.0 &	19.1 &	101.7 &	54.5 &	32.3 &	16.5 &	103.3 &	6.1 &	3.0 & 2.9 \\
c6288&	1.5 &	928.3&	60.5 &	988.8 &	23.7 &	57.4 &	81.1 &	51.2 &	38.7 &	61.6 &	151.5 &	56.4 &	23.5 &	70.5 &	150.3 &	12.2 &	6.5 &	6.6 \\
c7552&	0.5 &	358.8 &	17.0 &	375.8	& 29.3 & 18.3 &	47.6 &	90.6 &	39.2 &	21.1 &	150.9 &	99.6 &	23.2 &	17.7 &	140.5 &	7.9 &	2.5 &	2.7 \\
dalu&	0.4 &	267.7 &	9.2&	276.9 &	27.7&	9.2&	36.9&	74.5&	40.0 &	13.9&	128.4 &	82.0 &	43.1 &	12.7&	137.7 &	7.5&	2.2&	2.0 \\
i10&	0.5 &	342.7 &	21.4 &	364.0 &	33.7 &	17.0 &	50.7 &	98.6 &	35.2 &	20.5&	154.3 &	108.4 &	40.4&	23.2&	172.0 &	7.2&	2.4&	2.1 \\
k2&	0.3&	308.7&	13.6&	322.3&	38.1&	12.1&	50.2&	84.5&	35.7&	14.8&	134.9&	92.9 &	62.2&	15.1&	170.2 &	6.4&	2.4&	1.9 \\
mainpla&	0.8&	636.7&	47.9&	684.6&	30.2&	46.4&	76.6&	99.4&	30.7&	49.2&	179.2&	109.3 &	29.8&	45.8&	184.9 &	8.9&	3.8&	3.7 \\
div &	12.7&	6554.4&	291.9&	6846.2&	115.6&	250.2&	365.8&	104.7&	171.7&	280.2&	556.6&	115.2 &	379.4&	309.5&	804.1 &	18.7 &	12.3&	8.5 \\
log2 &	11.0 &	7614.2&	587.5&	8201.7	&67.5&	409.4 &	476.9	&110.6	&70.3&	569.4 &	750.3&	121.6 &	165.0&	571.0 &	857.6 &	17.2&	10.9&	9.6 \\
max&	0.7&	397.9&	22.5&	420.4&	30.2&	17.0&	47.2&	78.7&	30.2&	23.4&	132.3&	86.5 &	27.3&	25.3&	139.2 &	8.9&	3.2&	3.0 \\
multiplier&	8.2&	5359.1&	487.2&	5846.3&	58.7&	394.8&	453.5&	97.9&	117.9 &	457.4&	673.2&	107.7 &	58.4&	408.2&	574.3 &	12.9&	8.7&	10.2 \\
sin&	1.9&	1216.4&	342.6&	1559.0&	63.4&	85.6&	149.0&	52.4&	34.3&	110.8&	197.5&	57.6 &	26.5&	89.8&	173.9	&10.5&	7.9&	9.0 \\
sqrt&	7.4&	4460.3&	340.6&	4800.9 &	39.9&	294.1&	334.0 &	110.5 &	61.7 &	340.0 &	512.2 &	121.5 &	55.1 &	346.6&	523.2 &	14.4&	9.4&	9.2 \\
square&	6.7 &	3913.0 &	345.6&	4258.6&	58.7&	313.8&	372.5&	100.4&	23.5&	360.9&	484.7&	110.4  &44.6&	273.3&	428.3 &	11.4&	8.8&	10.0 \\
aes\_secworks&	6.2&	5526.7	&310.5&	5837.2&	77.2&	282.4&	359.6&	101.2&	138.8&	333.8	&573.9&	111.4	&94.4&	348.2&	553.9 &	16.2&	10.2&	10.5 \\
aes\_xcrypt	&9.0 &	8293.0	&489.8&	8782.8	&201.1&	462.2&	663.3&	99.9&	128.9&	504.7&	733.5&	109.9 &	296.3&	457.9&	864.0 &	13.2&	12.0&	10.2 \\
bp\_be&	18.9&	14890.4&	1457.9&	16348.2&	219.5&	1212.3&	1431.8	&145.6	&503.5 &	1523.9&	2172.9	&160.1 &	396.3&	1247.8&	1804.2 &	11.4&	7.5&	9.1 \\
ethernet&	8.26&	7014.6&	514.2&	7528.9&	301.1&	454.7&	755.8&	157.0&	278.0&	525.6&	960.6&	172.7 &	226.9&	509.8&	909.4 &	10.0&	7.8&	8.3 \\
jpeg&	39.7&	27934.4&	2220.2&	30154.7&	257.4&	2154.1&	2411.4&	175.5&	480.8&	2231.6&	2887.9&	193.1 &	515.6&	2147.9&	2856.6 &	12.5&	10.4&	10.6 \\
picosoc	&10.6&	8714.8&	614.0&	9328.8&	273.6&	583.2&	856.9&	112.4&	462.4&	623.6&	1198.3&	123.6	&587.4	&567.5	&1278.5 &	10.9&	7.8&	7.3 \\
tinyRocket&	8.1&	7366.6	&536.6&	7903.3	&195.4&	517.2&	712.6&	101.2&	310.4&	554.1&	965.7&	111.4 &	154.5&	524.7&	790.6 &	11.1&	8.2&	10.0 \\
vga\_lcd&	15.6&	13397.8&	1101.2	&14498.9&	256.3&	923.8&	1180.0&	180.6&	410.1&	1079.1&	1669.8&	198.6	&412.2&	973.2	&1584.0 &	12.3&	8.7&	9.2\\
wb\_conmax	&6.6 &	6496.6&	469.9&	6966.5	&103.8&	465.2&	569.0&	82.0&	301.2&	417.2&	800.5&	90.2 &	327.6&	436.1	&853.9 &	12.2&	8.7&	8.2\\
\bottomrule
\end{tabular}
}
\end{table*}

\subsubsection{Analysis of Results} We compare \ac{QoR} of synthesis output using \solution{} generated recipes to our baselines and \ac{SOTA} prior work~\cite{mlcad_abc,iccad_2021} in \autoref{tab:bullseye_sota}. 
This shows synthesis output using \solution{} generated recipes achieve better \ac{QoR} compared to conventional \ac{SA} and resyn2$^*$ with \textbf{2.94x} average runtime speedup. 
\rtbf{Red} and \btbf{Blue} denotes top-$2$ performing recipes.
Our results show \solution{} generated recipes using finetuning perform better than \ac{SOTA} on $9$ out of $10$ benchmarks (~\autoref{tab:bullseye_sota}). 
Compared to \cite{mlcad_abc} and \cite{iccad_2021}, our best predictor ($FT+A$) achieves better \textit{rf \%} (around \textbf{4\%}) with a \textbf{11.20x} and \textbf{1.04x} run-time speed-up, respectively. 
Our ($FT+R$) predictor achieves better \textit{rf \%} than \cite{mlcad_abc} and \cite{iccad_2021} (\textbf{2.7\%} and \textbf{1.35\%}) with \textbf{12x} and \textbf{1.10x} run-time speed-up, respectively. The zero-shot predictor $ZS$ obtains almost similar \ac{QoR}, however achieving a huge run-time speed-up over prior work (\textbf{45x} and \textbf{4x} compared to \cite{mlcad_abc,iccad_2021} respectively). This shows that our zero-shot predictor is indeed helpful for prediction on unseen netlists. The prediction quality improved through fine-tuning leads to better recipes, i.e., recipes found using the active learning guided fine-tuned ($FT+A$) model perform better compared to those found by the Zero-shot ($ZS$) and random recipe augmented fine-tuned ($FT+R$) \ac{QoR} predictors.

\begin{figure*}[t]
\centering
     
	\subfloat[bp\_be]{
		\begin{tikzpicture}[scale=0.3] 
		\begin{axis}
			[    xmode=log,
            log ticks with fixed point,
			height=0.8\columnwidth,
			width=1.4\columnwidth,
			xlabel= Epsilon ($\epsilon$), 
			ylabel= Accuracy,
			every major tick/.append style={very thick, major tick length=10pt, black},
			axis line style = very thick,
			                    tick label style={font=\Huge},
			                                        label style={font=\Huge},
			                    xlabel=\textbf{Time(s)},ylabel=\textbf{\# Nodes},
			grid style = dashed,
			grid=both,
			legend style=
			{at={(0.8,0.8)}, 
				anchor=south west, 
				anchor= north , 
			} ,
			]
			\addplot[color=blue,solid,mark=*,line width=1] plot coordinates {
				(0.5,82514)
				(1433.63,78186)
				(14890.36,78012)
			};

			\addplot[color=olive,solid,mark=*,line width=1] plot coordinates {
                (0.5,82514)
                (2.1,78595)
                (34.97,78487)
                (97.31,78446)
                (219.45,78262)
			};

			\addplot[color=red,solid,mark=*,line width=1] plot coordinates {
                (0.5,82514)
                (100,82514)
                (102.9,78467)
                (150.01,78407)
                (314.35,78064)
                (496.33,78064)
			};

		\legend{\Huge{\bf{B}}\\\Huge{\bf{ZS}}\\\Huge{\bf{$F^{c}$}}\\}
			\end{axis}
		\end{tikzpicture}
	}
	\subfloat[ethernet]{
		\begin{tikzpicture}[scale=0.3] 
		\begin{axis}
			[    xmode=log,
            log ticks with fixed point,
			height=0.8\columnwidth,
			width=1.4\columnwidth,
			every major tick/.append style={very thick, major tick length=10pt, black},
			axis line style = very thick,
			                    tick label style={font=\Huge},
			                                        label style={font=\Huge},
			                    xlabel=\textbf{Time(s)},ylabel=\textbf{\# Nodes},
			grid style = dashed,
			grid=both,
			legend style=
			{at={(0.8,0.8)}, 
				anchor=south west, 
				anchor= north , 
			} ,
			]
			\addplot[color=blue,solid,mark=*,line width=1] plot coordinates {
				(0.5,    67164)
                (711.19,64844)
                (7014.61,64821)
			};

			\addplot[color=olive,solid,mark=*,line width=1] plot coordinates {
				(0.5,    67164)
                (1.83, 65011)
                (19.1, 65000)
                (110.13,64947)
                (301.08, 64968)
                (415.15, 64945)
			};

			\addplot[color=red,solid,mark=*,line width=1] plot coordinates {
			    (0.5,    67164)
			    (100,    67164)
                (101.85, 64920)
                (117.66, 64863)
                (214.76, 64870)
                (326.93,64858)
			};

		\legend{\Huge{\bf{B}}\\\Huge{\bf{ZS}}\\\Huge{\bf{$F^c$}}\\}
			\end{axis}
		\end{tikzpicture}
	}
  \subfloat[jpeg]{
		\begin{tikzpicture}[scale=0.3] 
		\begin{axis}
			[    xmode=log,
            log ticks with fixed point,
			height=0.8\columnwidth,
			width=1.4\columnwidth,
			xlabel= Epsilon ($\epsilon$), 
			ylabel= Accuracy,
			every major tick/.append style={very thick, major tick length=10pt, black},
			axis line style = very thick,
			                    tick label style={font=\Huge},
			                                        label style={font=\Huge},
			                    xlabel=\textbf{Time(s)},ylabel=\textbf{\# Nodes},
			grid style = dashed,
			grid=both,
			legend style=
			{at={(0.8,0.8)}, 
				anchor=south west, 
				anchor= north , 
			} ,
			]
			\addplot[color=blue,solid,mark=*,line width=1] plot coordinates {
				(0.5,114771)
				(2938.01, 102441)
				(27934.44, 102341 )
			};

			\addplot[color=olive,solid,mark=*,line width=1] plot coordinates {
				(0.5,114771)
				(4.71, 103019)
				(19.28, 102922)
				(97.1, 102863)
				(257.35,102616)
			};

			\addplot[color=red,solid,mark=*,line width=1] plot coordinates {
			    (0.5,114771)
			    (100,114771)
			    (103.34,103196)
			    (149.38,102558)
			    (360.41,102441)
			    (515.64,102421)
			};

		\legend{\Huge{\bf{B}}\\\Huge{\bf{ZS}}\\\Huge{\bf{$F^c$}}\\}
			\end{axis}
		\end{tikzpicture}
	}
	\subfloat[picosoc]{
		\begin{tikzpicture}[scale=0.3] 
		\begin{axis}
			[    xmode=log,
            log ticks with fixed point,
			height=0.8\columnwidth,
			width=1.4\columnwidth,
			xlabel= Epsilon ($\epsilon$), 
			ylabel= Accuracy,
			every major tick/.append style={very thick, major tick length=10pt, black},
			axis line style = very thick,
			                    tick label style={font=\Huge},
			                                        label style={font=\Huge},
			                    xlabel=\textbf{Time(s)},ylabel=\textbf{\# Nodes},
			grid style = dashed,
			grid=both,
			legend style=
			{at={(0.8,0.8)}, 
				anchor=south west, 
				anchor= north , 
			} ,
			]
			\addplot[color=blue,solid,mark=*,line width=1] plot coordinates {
				(0.5,  82945)
                (866.8, 76515)
                (8714.8, 76490)
			};

			\addplot[color=olive,solid,mark=*,line width=1] plot coordinates {
				(0.5,  82945)
				(2.14,77410)
				(47.98,76738)
				(101.23,76704)
				(273.64,76698)
			};

			\addplot[color=red,solid,mark=*,line width=1] plot coordinates {
			    (0.5,  82945)
			    (100,  82945)
                (104.34, 76673)
                (127.67, 76753)
                (417.93, 76612)
                (687.41, 76598)
			};

		\legend{\Huge{\bf{B}}\\\Huge{\bf{ZS}}\\\Huge{\bf{$F^c$}}\\}
			\end{axis}
		\end{tikzpicture}
	}
\vspace{0.2in}
	\subfloat[tinyRocket]{
		\begin{tikzpicture}[scale=0.3] 
		\begin{axis}
			[    xmode=log,
            log ticks with fixed point,
			height=0.8\columnwidth,
			width=1.4\columnwidth,
			xlabel= Epsilon ($\epsilon$), 
			ylabel= Accuracy,
			every major tick/.append style={very thick, major tick length=10pt, black},
			axis line style = very thick,
			                    tick label style={font=\Huge},
			                                        label style={font=\Huge},
			                    xlabel=\textbf{Time(s)},ylabel=\textbf{\# Nodes},
			grid style = dashed,
			grid=both,
			legend style=
			{at={(0.8,0.8)}, 
				anchor=south west, 
				anchor= north , 
			} ,
			]
			\addplot[color=blue,solid,mark=*,line width=1] plot coordinates {
				(0.5,52315)
				(707.28,40793)
				(7366.62,40717)
			};

			\addplot[color=olive,solid,mark=*,line width=1] plot coordinates {
			    (0.5,52315)
				(1.46,40868)
				(17.12,40835)
				(97.16,40835)
				(195.4,40820)
			};

			\addplot[color=red,solid,mark=*,line width=1] plot coordinates {
			  (0.5,52315)
              (100,52315)
			  (101.4,41061)
			  (112.1,40961)
			  (160.14,40954)
			  (254.54,40890)
			};

		\legend{\Huge{\bf{B}}\\\Huge{\bf{ZS}}\\\Huge{\bf{$F^c$}}\\}
			\end{axis}
		\end{tikzpicture}
	}
\subfloat[vga\_lcd]{
		\begin{tikzpicture}[scale=0.3] 
		\begin{axis}
			[    xmode=log,
            log ticks with fixed point,
			height=0.8\columnwidth,
			width=1.4\columnwidth,
			every major tick/.append style={very thick, major tick length=10pt, black},
			axis line style = very thick,
			                    tick label style={font=\Huge},
			                                        label style={font=\Huge},
			                    xlabel=\textbf{Time(s)},ylabel=\textbf{\# Nodes},
			grid style = dashed,
			grid=both,
			legend style=
			{at={(0.8,0.8)}, 
				anchor=south west, 
				anchor= north , 
			} ,
			]
			\addplot[color=blue,solid,mark=*,line width=1] plot coordinates {
			   (0.5,105334)
			   (1314.3,103539)
			   (13397.77,103509)
			};

			\addplot[color=olive,solid,mark=*,line width=1] plot coordinates {
			    (0.5,105334)
                (2.69,103637)
                (25.35,103668)
                (126.96,103629)
                (256.25,103647)
			};

			\addplot[color=red,solid,mark=*,line width=1] plot coordinates {
			   (0.5,105334)
			   (100,105334)
			   (102.72,103633)
			   (132.81,103679)
			   (329.24,103626)
			   (512.24,103601)
			};

		\legend{\Huge{\bf{B}}\\\Huge{\bf{ZS}}\\\Huge{\bf{$F^c$}}\\}
			\end{axis}
		\end{tikzpicture}
	}
  \subfloat[wb\_conmax]{
		\begin{tikzpicture}[scale=0.3] 
		\begin{axis}
			[    xmode=log,
            log ticks with fixed point,
			height=0.8\columnwidth,
			width=1.4\columnwidth,
			xlabel= Epsilon ($\epsilon$), 
			ylabel= Accuracy,
			every major tick/.append style={very thick, major tick length=10pt, black},
			axis line style = very thick,
			                    tick label style={font=\Huge},
			                                        label style={font=\Huge},
			                    xlabel=\textbf{Time(s)},ylabel=\textbf{\# Nodes},
			grid style = dashed,
			grid=both,
			legend style=
			{at={(0.8,0.8)}, 
				anchor=south west, 
				anchor= north , 
			} ,
			]
			\addplot[color=blue,solid,mark=*,line width=1] plot coordinates {
			    (0.5,47840)
			    (649.52,43910)
			    (6496.56,43859)
			};

			\addplot[color=olive,solid,mark=*,line width=1] plot coordinates {
                (0.5,47840)
			    (1.21,44662)
			    (10.36,44679)
			    (51.586,44090)
			    (103.8,43997)
			};

			\addplot[color=red,solid,mark=*,line width=1] plot coordinates {
                (0.5,47840)
                (100,47840)
			    (102.74,44001)
			    (117.71,43926)
			    (316.09,43924)
			    (427.58,43876)
			};

		\legend{\Huge{\bf{B}}\\\Huge{\bf{ZS}}\\\Huge{\bf{$F^c$}}\\}
			\end{axis}
		\end{tikzpicture}
	}
	\subfloat[multiplier]{
		\begin{tikzpicture}[scale=0.3] 
		\begin{axis}
			[    xmode=log,
            log ticks with fixed point,
			height=0.8\columnwidth,
			width=1.4\columnwidth,
			ylabel= Accuracy,
			every major tick/.append style={very thick, major tick length=10pt, black},
			axis line style = very thick,
			                    tick label style={font=\Huge},
			                                        label style={font=\Huge},
			                    xlabel=\textbf{Time(s)},ylabel=\textbf{\# Nodes},
			grid style = dashed,
			grid=both,
			legend style=
			{at={(0.8,0.8)}, 
				anchor=south west, 
				anchor= north , 
			} ,
			]
			\addplot[color=blue,solid,mark=*,line width=1] plot coordinates {
			        (0.5,27062)
			        (5359.1,24362)
			        (24400.8,24360)
			};
			
			\addplot[color=olive,solid,mark=*,line width=1] plot coordinates {
			        (0.5,27062)
			        (1,24585)
			        (9.3,24542)
			        (29.6,24542)
			        (63.4,24541)
			};

			\addplot[color=red,solid,mark=*,line width=1] plot coordinates {
			     (0.5,27062)
			     (100, 27062)
			     (100.7,24641)
			     (105.9,24641)
			     (129.1,24513)
			     (158.4,24368)
			};

		\legend{\Huge{\bf{B}}\\\Huge{\bf{ZS}}\\\Huge{\bf{$F^c$}}\\}
			\end{axis}
		\end{tikzpicture}
	}
	\vspace{0.3in}
    \caption{\ac{QoR} at various times from recipes by simulated annealing using different evaluators: actual synthesis ($B$) and proxy predictors ($ZS$ and $FT+A$)}
    \label{fig:frameworkTime}
\end{figure*}
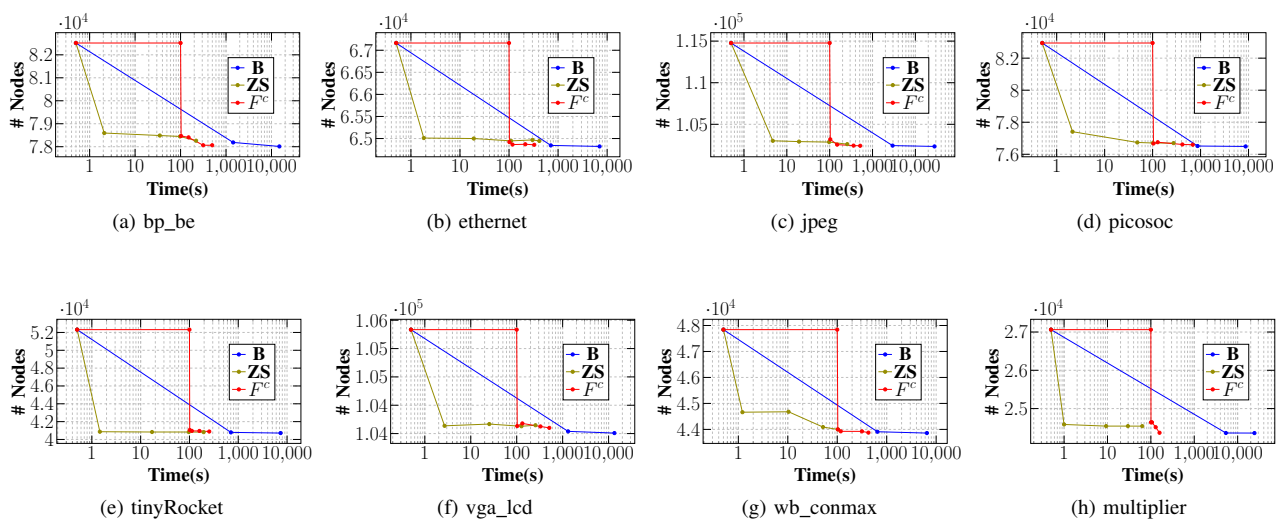

\autoref{tab:bullseye_large} shows the effectiveness of \solution{} on \textbf{large} benchmarks ($\geq 40k$ nodes) where \textbf{\ac{SOTA} prior works do not work}. We compared the \ac{QoR} of synthesis output generated using recipes from \ac{SA} $+$ \solution{} as the back-end evaluator with our baselines. 
The results indicate high \textit{RI} with respect to resyn2$^*$ in reduction factor using recipes generated from \solution{} predictors. \autoref{tab:bullseye_time} shows runtime complexity and speedup of \solution{} guided recipe generation. We present relative improvement and timing speedup of \solution{} in \autoref{fig:bullseyeQualitative}. We observe that $FT+A$-generated recipes achieve similar (and in one case significantly better) \ac{QoR} with \textbf{$\geq$ 2x} run-time speed-up over $SA_{1k}$. $div$ achieved $>100 \%$ \textit{RI} compared to resyn2$^*$ indicating that \solution{} generated recipe optimized twice the number of nodes compared to resyn2 applied for the same time duration. Another crucial observation is: runtime speedup improves with more number of nodes in initial \ac{AIG}. All designs having $\geq 10k$ nodes in initial \ac{AIG} achieved more than \textbf{8x} timing speed-up with similar \ac{QoR} quality compared to $SA_{1k}$. To illustrate the faster recipe space search that comes from using \solution{} \ac{QoR} predictors, we chart the \ac{QoR} progression over time in ~\autoref{fig:frameworkTime}. The graphs show \solution{} scales on large benchmarks and can generate superior quality synthesis recipes quickly compared to the conventional \ac{SA} approach that uses actual synthesis runs. Note that the graph for $FT+A$ includes an initial time penalty from synthesis data generation and fine-tuning the model; after this the models enable very quick exploration of the recipe space due to fast inference by the proxy models. 

\section{Discussion and Insights}
\label{sec:discuss}

We proposed \textbf{\solution{}} as a solution for \ac{ML}-guided synthesis recipe generation that scales on large designs. 
\solution{} uses few-shot learning and active learning to build upon models pre-trained on historical synthesis data, particularly given a limited synthesis run budget for new, unseen designs. 
Our framework generates better synthesis recipes compared to prior techniques, with an average $2.94$x run-time speed-up and up to $10.2$x speed-up (on large benchmarks) compared to conventional black-box approaches.
We have made our code and associated dataset available to the community at: \url{https://github.com/NYU-MLDA/OpenABC}.


\bibliographystyle{IEEEtran}
\bibliography{main.bib}
\end{document}